\documentclass[journal]{IEEEtran} 

\usepackage{algorithmic}

\usepackage{algorithm}
\usepackage{color}
\usepackage{graphicx}
\usepackage[table]{xcolor}

\usepackage{booktabs}
\usepackage{multirow}
\usepackage{gensymb}
\usepackage{amsmath}
\usepackage{CJKutf8}
\usepackage{subcaption}

\usepackage{threeparttable}

\usepackage{booktabs}

\usepackage{float}

\usepackage{amsfonts,amssymb}

\usepackage[
    backend=biber,
    sorting=none,
    url=false,
    isbn=false,
    doi=false
]{biblatex}

\newcolumntype{L}[1]{>{\raggedright\arraybackslash}p{#1}}
\newcolumntype{C}[1]{>{\centering\arraybackslash}p{#1}}
\newcolumntype{R}[1]{>{\raggedleft\arraybackslash}p{#1}}
 
\IEEEoverridecommandlockouts                              

\title{\LARGE \bf
Which LiDAR scanning pattern is better for roadside perception: Repetitive or Non-repetitive?
}

\author{Zhiqi Qi, Runxin Zhao, Hanyang Zhuang, Member, IEEE, \\
Chunxiang Wang, Member, IEEE, and Ming Yang, Member, IEEE
\thanks{This work was supported by National Natural Science Foundation of China (62203294/U22A20100/62373250). Hanyang Zhuang and Ming Yang are the corresponding authors.}
\thanks{Zhiqi Qi, Runxin Zhao, Chunxiang Wang, and Ming Yang
 are with the School of Automation and Intelligent Sensing, Shanghai Jiao Tong University, Shanghai, 200240, China; Key Laboratory of System Control and Information Processing, Ministry of Education of China, Shanghai, 200240, China (email: mingyang@sjtu.edu.cn).}%
\thanks{Hanyang Zhuang is with Global College, Shanghai Jiao Tong University (zhuanghany11@sjtu.edu.cn).} }%

\begin{document}
\begin{CJK}{UTF8}{gbsn} 

\maketitle
\thispagestyle{empty}
\pagestyle{empty}

\begin{abstract}
LiDAR-based roadside perception is a cornerstone of advanced Intelligent Transportation Systems (ITS). While considerable research has addressed optimal LiDAR placement for infrastructure, the profound impact of differing LiDAR scanning patterns on perceptual performance remains comparatively under-investigated. The inherent nature of various scanning modes—such as traditional repetitive (mechanical/solid-state) versus emerging non-repetitive (e.g., prism-based) systems—leads to distinct point cloud distributions at varying distances, critically dictating the efficacy of object detection and overall environmental understanding. To systematically investigate these differences in infrastructure-based contexts, we introduce the “InfraLiDARs’ Benchmark,” a novel dataset meticulously collected in the CARLA simulation environment using concurrently operating infrastructure-based LiDARs exhibiting both scanning paradigms. Leveraging this benchmark, we conduct a comprehensive statistical analysis of the respective LiDAR scanning abilities and evaluate the impact of these distinct patterns on the performance of various leading 3D object detection algorithms. Our findings reveal that non-repetitive scanning LiDAR and the 128-line repetitive LiDAR were found to exhibit comparable detection performance across various scenarios. Despite non-repetitive LiDAR’s limited perception range, it’s a cost-effective  option considering its low price. Ultimately, this study provides insights for setting up roadside perception system with optimal LiDAR scanning patterns and compatible algorithms for diverse roadside applications, and publicly releases the "InfraLiDARs’ Benchmark” dataset to foster further research.
\end{abstract}


\section{Introduction}
\begin{figure}[htbp] 
    \centering 
    \includegraphics[width=1\linewidth]{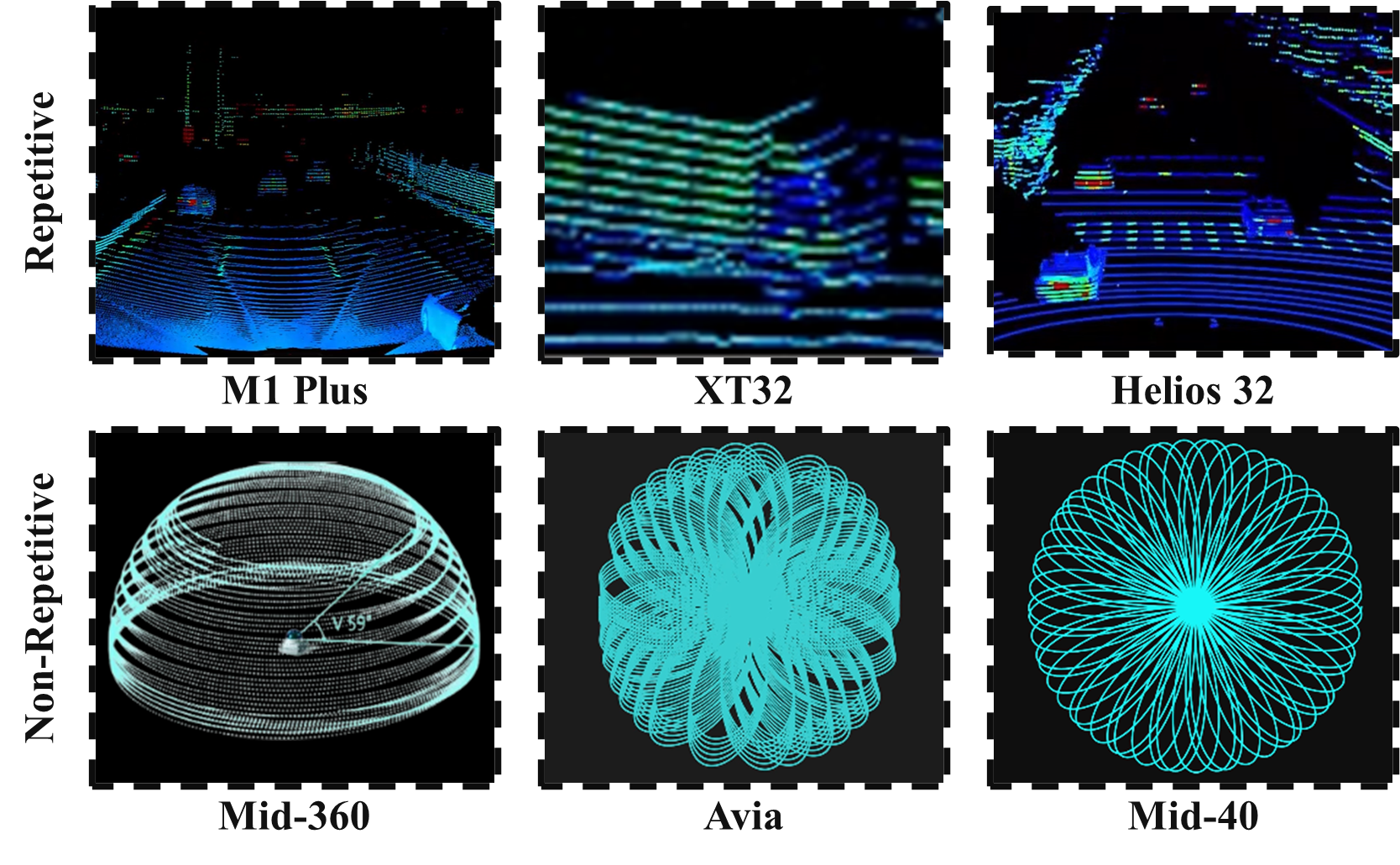}
    \caption{Visualization of different LiDAR point clouds. The three images in the upper section depict point cloud patterns from repetitive scanning LiDARs, while those in the lower section depict patterns from non-repetitive scanning LiDARs.}
    \label{fig:different_patterns_of_pointcloud_visualization} 
\end{figure}

\IEEEPARstart{L}{iDAR} technology deployed on roadside infrastructure is instrumental in advancing Intelligent Transportation Systems (ITS) and the broader vision of smart cities, delivering precise, real-time three-dimensional data crucial for comprehensive environmental perception that underpins enhanced traffic safety, optimized flow management, improved urban mobility, and cooperative autonomous driving capabilities. Traditional approaches have often relied on mechanical or solid-state LiDARs with repetitive scanning patterns for such fixed-position deployments. These LiDARs operate with a predetermined, fixed N-line pattern, thereby repeatedly surveying the same spatial regions in each operational cycle. Conversely, non-repetitive scanning LiDARs (such as the Livox Avia or Mid-series) employ dynamically evolving scan patterns designed to progressively densify point cloud coverage across the entire field of view with increasing integration time. Typical point clouds from two distinct scanning patterns of LiDAR systems are depicted in Figure \ref{fig:different_patterns_of_pointcloud_visualization}.

Optimal roadside LiDAR configuration hinges on both sensor placement and its intrinsic scanning pattern. While LiDAR placement strategies for infrastructure have been extensively studied \cite{10950411}\cite{10377313}, the comparative performance implications of different scanning patterns—such as the trade-offs between fixed-path repetitive scanning and dynamically evolving non-repetitive scanning—remain largely unaddressed in these fixed contexts. This study directly tackles this critical gap by systematically evaluating these diverse scanning modalities to elucidate their respective strengths and weaknesses for infrastructure-based perception tasks.

This paper, leveraging a specialized simulation framework developed within the CARLA environment and introducing the "InfraLiDARs' Benchmark", a novel, comprehensive dataset—aims to systematically investigate the multifaceted factors influencing infrastructure-based LiDAR perception. The dataset is available online\footnote{\url{https://www.kaggle.com/datasets/zhiqiqi/infralidars-benchmark/}}. Our research conducts a rigorous comparative analysis of the impact of diverse LiDAR scanning patterns, varying traffic scenarios (including urban crossroads, highways, and curves), and a range of 3D object detection algorithms on overall system performance for tasks such as vehicle detection and localization. Through this extensive empirical evaluation, the study ultimately seeks to provide actionable, data-driven recommendations for optimal LiDAR scanning mode configurations tailored to specific roadside application requirements and operational contexts.

The primary contributions of this paper are threefold:
\begin{itemize}
    \item 
    Develop an analytical framework to assess the detection effectiveness of diverse LiDAR systems, comprising both a statistical benchmark of scanning patterns and a performance benchmark of perception algorithms.

    \item 
    Systematically benchmark the compatibility of diverse LiDAR scanning patterns with various algorithms, yielding foundational knowledge crucial for determining optimal algorithm choices for LiDAR systems.

    \item 
    Publicly release a novel, extensive dataset tailored for roadside LiDAR perception research, providing data across various scenarios from multiple LiDARs with distinct scanning patterns.

\end{itemize}

\section{Related Works}
\subsection{Researches on LiDAR Scanning Patterns}
Prior LiDAR research commonly employs a classification framework \cite{kim2021nanophotonics} that distinguishes three main types: mechanical, solid-state, and hybrid solid-state. Within this taxonomy, the solid-state designation primarily refers to technologies like flash LiDAR \cite{lemmetti2021long} and optical phased array (OPA)-based LiDAR \cite{poulton2019long}, while microelectromechanical systems (MEMS)-based LiDARs \cite{wang2018compact} are typically categorized under hybrid solid-state systems. These patterns directly influence critical attributes of the resulting point cloud, such as the Field of View (FoV) coverage, the point density achieved on objects at different distances and orientations, the uniformity of point distribution across the scene, and the rate at which information about the environment is accumulated over successive scans.

Our research utilizes a LiDAR classification scheme predicated on scanning patterns, specifically differentiating between repetitive and non-repetitive operational modes\cite{LivoxWiki}. Research has shown that the non-repetitive scanning patterns of LiDAR could provide a much higher resolution than conventional LiDARs and feature a peaked central angular density\cite{9351673}. Furthermore, our pattern-focused methodology, distinct from conventional principle-based categorizations, is essential for evaluating their performance efficacy in diverse downstream tasks and for generating valuable insights applicable to future research and production endeavors.

\subsection{3D Object Detection}
The field of 3D Object Detection using 3D point clouds predominantly relies on deep neural network methodologies. Within this context, our research narrows its focus to the specific application of evaluating vehicle detection efficacy using algorithms tailored for point clouds from outdoor, infrastructure-based LiDAR. This investigation aims to facilitate effective vehicle tracking and localization in such fixed-sensor environments.

Early approaches to 3D object detection from point clouds included point-based methods, which operate directly on the raw, unstructured data\cite{qi2017pointnetdeeplearningpoint}\cite{qi2017pointnetdeephierarchicalfeature}\cite{shi2019pointrcnn3dobjectproposal}\cite{yang2019stdsparsetodense3dobject}\cite{shi2020pointgnngraphneuralnetwork}. The pioneering PointNet\cite{qi2017pointnetdeeplearningpoint}, for instance, applied Multilayer Perceptrons (MLPs) to individual points but struggled with capturing local contextual information. 

Subsequently, voxel-based methods gained prominence by discretizing the point cloud into a regular 3D voxel grid, allowing for the application of standard 3D convolutional networks for feature encoding\cite{zhou2017voxelnetendtoendlearningpoint}\cite{s18103337}\cite{shi2020pointsparts3dobject}\cite{deng2021voxelrcnnhighperformance}. A key advancement in this area was SECOND (Sparsely Embedded Convolutional Detection)\cite{s18103337}, which significantly enhanced processing efficiency by employing sparse convolutions, which selectively perform computations only on non-empty voxels.To further accelerate detection, pillar-based methods\cite{lang2019pointpillarsfastencodersobject}\cite{shi2022pillarnetrealtimehighperformancepillarbased} like PointPillars\cite{lang2019pointpillarsfastencodersobject} were introduced. This approach partitions the point cloud into pillars across the ground plane. Features are learned within each pillar, enabling fast and efficient object detection using 2D convolutional networks.

More recently, Transformer-based methods have shown significant promise by leveraging attention mechanisms to model complex relationships within point cloud data.\cite{wang2021detr3d3dobjectdetection}\cite{wang2023dsvtdynamicsparsevoxel}\cite{sheng2021improving3dobjectdetection}. DSVT\cite{wang2023dsvtdynamicsparsevoxel}, for example, employs a dynamic and fully sparse Transformer backbone, utilizing a novel rotated set attention mechanism on voxel features for efficient and accurate object recognition.

Several open-source platforms consolidate multiple 3D object detection algorithms, such as MMDetection3D\cite{mmdetection3d2020} and extensions for Detectron2\cite{wu2019detectron2}. Among these, we selected OpenPCDet\cite{openpcdet2020} for this work. OpenPCDet stands out due to its highly modular design and comprehensive support for a diverse range of state-of-the-art 3D detection algorithms. These features streamline the development, comparison, and evaluation process, making it an effective choice for our research.

\subsection{Infrastructure-based LiDAR Dataset}

Infrastructure-based Lidar has numerous impactful applications in advancing ITS. Zhengwei Bai et al. review infrastructure-based object detection and tracking for Cooperative Driving Automation, highlighting how roadside perception system can significantly enhance the perception capabilities of connected vehicles by overcoming the inherent range and occlusion limitations of onboard sensors\cite{bai2022infrastructure}. George R et al. explore vehicle-to-infrastructure (V2I) cooperative sensing by defining requirements for its core components—information flow, sensing, perception, and mapping—and proposing optimized sensor suites and data processing strategies\cite{george2025infrastructure}. 

As for LiDAR dataset, a significant body of research and numerous publicly available datasets in LiDAR-based perception have predominantly centered on vehicle-mounted (ego-vehicle) configurations. Prominent examples that have catalyzed advancements in 3D object detection, tracking, and broader autonomous driving perception tasks include KITTI\cite{geiger2013vision}, the Waymo Open Dataset\cite{sun2020scalability}, nuScenes\cite{caesar2020nuscenes}, and Argoverse\cite{chang2019argoverse}. While these datasets are invaluable resources, they typically feature pre-defined sensor suites and their primary focus is not on providing a controlled, comparative analysis of the intrinsic perceptual differences arising from fundamentally distinct LiDAR sensor characteristics, such as varying line counts or scanning patterns (e.g., repetitive versus non-repetitive), under identical environmental conditions. A noteworthy recent study, emphasizing that "Non-Repetitive: A Promising LiDAR Scanning Pattern"\cite{xie2024non}, has commendably begun to address this by directly comparing different scanning modalities. However, this valuable research, akin to the vast majority of existing literature, maintains its focus on LiDAR systems mounted on vehicles. This leaves a crucial area underexplored: the comparative efficacy and optimal selection criteria for diverse LiDAR technologies and scanning patterns when they are deployed on roadside infrastructure for applications such as comprehensive traffic surveillance, precise vehicle localization, and cooperative perception, which is the central investigation of our work.

\section{Framework of benchmark}

\begin{figure}
    \centering
    \includegraphics[width=1\linewidth]{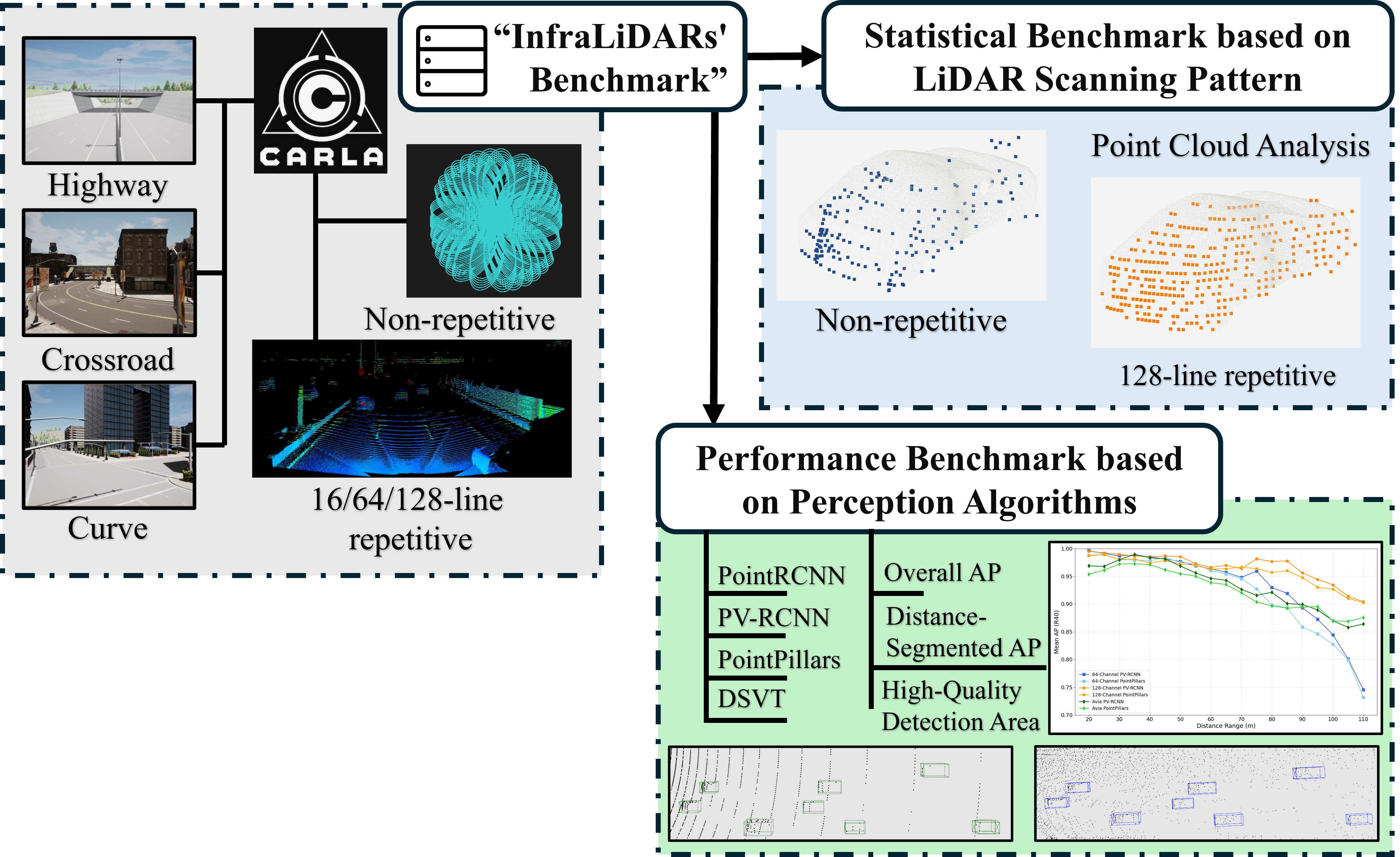}
    \caption{Framework of Benchmarking}
    \label{fig:benchmarking_framework}
\end{figure}

Our study introduces a comprehensive benchmarking framework, visually detailed in Figure \ref{fig:benchmarking_framework}, designed to systematically evaluate and compare LiDAR technologies and 3D object detection algorithms for infrastructure-based perception. This framework is anchored by our "InfraLiDARs' Benchmark" dataset, meticulously generated within the CARLA simulation environment. The dataset encompasses diverse traffic scenarios—specifically highway, crossroad, and curve environments—and incorporates data from a selection of LiDARs representing both multi-line repetitive (16, 64, and 128-line) and non-repetitive scanning patterns, reflecting distinct data acquisition approaches. The core of our benchmarking methodology comprises two primary analytical thrusts:

Firstly, a Statistical Benchmark based on Scanning Patterns is conducted. This involves a detailed Point Cloud Analysis to assess and contrast the intrinsic data acquisition characteristics resulting from different LiDAR types (such as the non-repetitive versus 128-line repetitive point clouds visually exemplified within Figure \ref{fig:benchmarking_framework}). Secondly, a Performance Benchmark based on Perception Algorithms is performed, where leading 3D object detection models (including PointRCNN, PV-RCNN, PointPillars and DSVT) are evaluated using metrics such as Overall AP, Distance-Segmented AP Analysis, and High-Quality Detection Area analysis.

The insights derived from these complementary benchmarks are then synthesized to inform Algorithm Selection and Analysis and to ultimately guide the proposal of optimized LiDAR and algorithm configurations tailored for various roadside applications.

\section{InfraLiDARs' Benchmark Dataset}

To generate point cloud data from diverse LiDAR types, encompassing both repetitive and non-repetitive scanning modes, we utilized the CARLA simulation environment. Opting for simulation was crucial because acquiring strictly comparable data from different LiDAR scanning modalities under identical real-world conditions is notably challenging, often complicated by calibration inconsistencies and environmental variability. Furthermore, simulation offers substantial advantages in terms of efficiency and cost-effectiveness for generating extensive datasets. For instance, ground truth labels are directly obtainable from the simulated environment, which significantly alleviates the data annotation burden and enhances label precision. Simulations also provide comprehensive environmental data often unavailable in real-world collection, such as a complete record of all objects within the sensor's purview; this facilitates nuanced analyses, including the proportion of detectable entities. Our choice of CARLA was primarily driven by its robust simulation engine, native support for multiple LiDAR scanning configurations, and its ability to conveniently generate diverse traffic participants. It is worth noting that similar benchmarking and validation objectives could likewise be achieved using other advanced simulation platforms, such as Gazebo.

\subsection{Scenarios}
\begin{figure}
    \centering
    \includegraphics[width=1\linewidth]{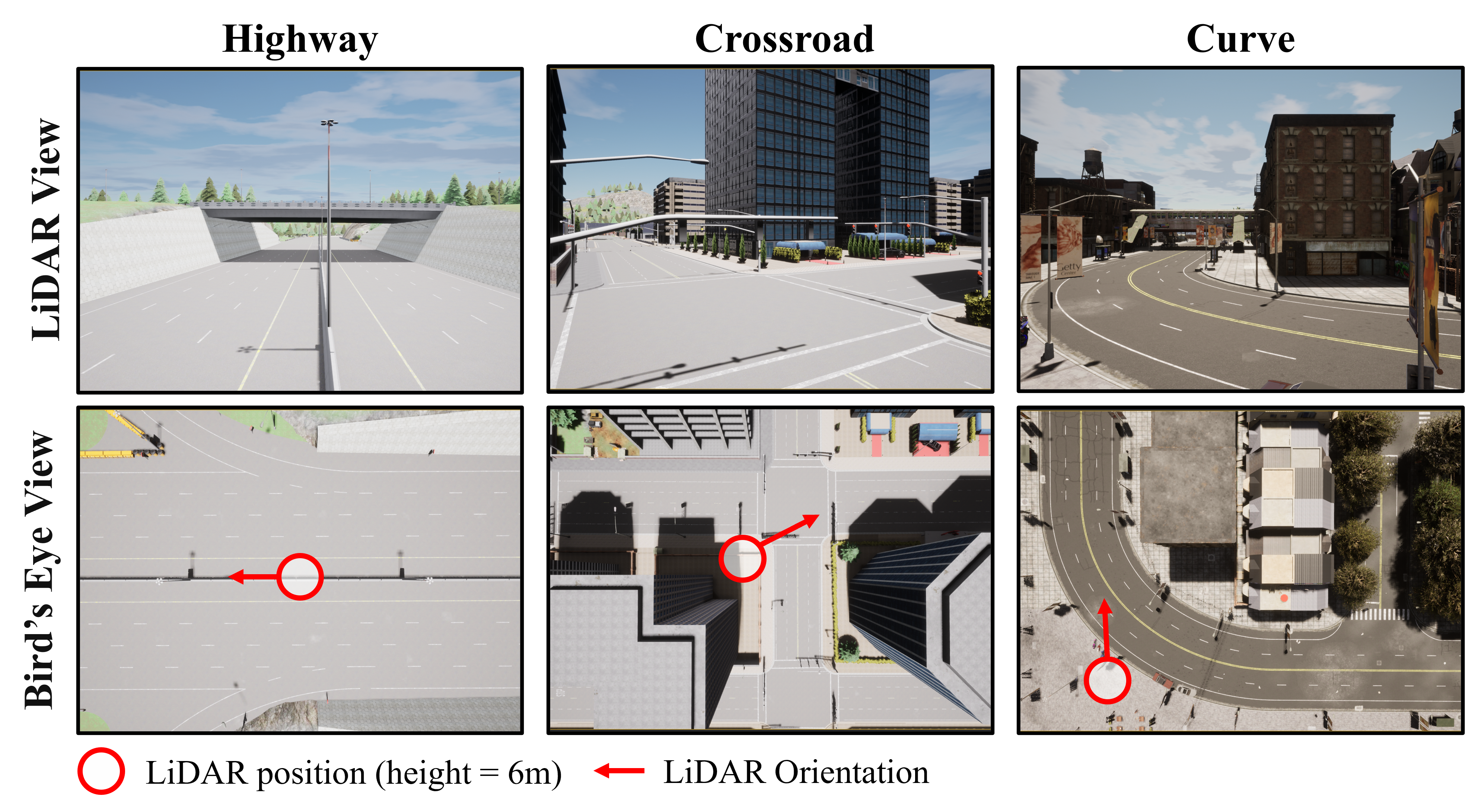}
    \caption{Lidar position and orientation}
    \label{Lidar position}
\end{figure}

\begin{figure}[htbp] 
    \centering 
    \begin{subfigure}[b]{0.48\textwidth} 
        \centering
        \includegraphics[width=\linewidth]{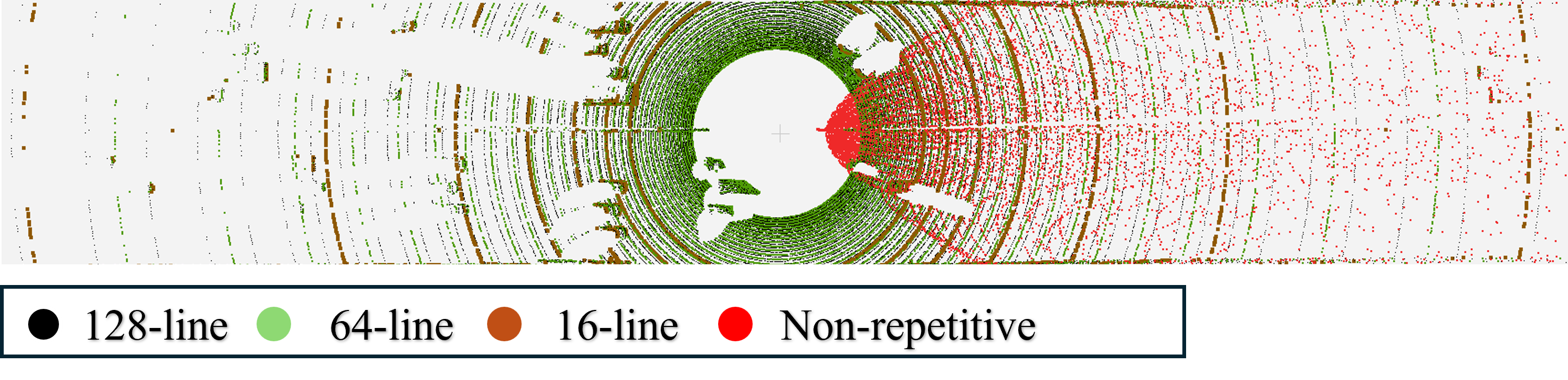} 
        \caption{Highway} 
        \label{fig:highway_pointcloud} 
    \end{subfigure}
    \begin{subfigure}[b]{0.2575\textwidth} 
        \centering
        \includegraphics[width=\linewidth]{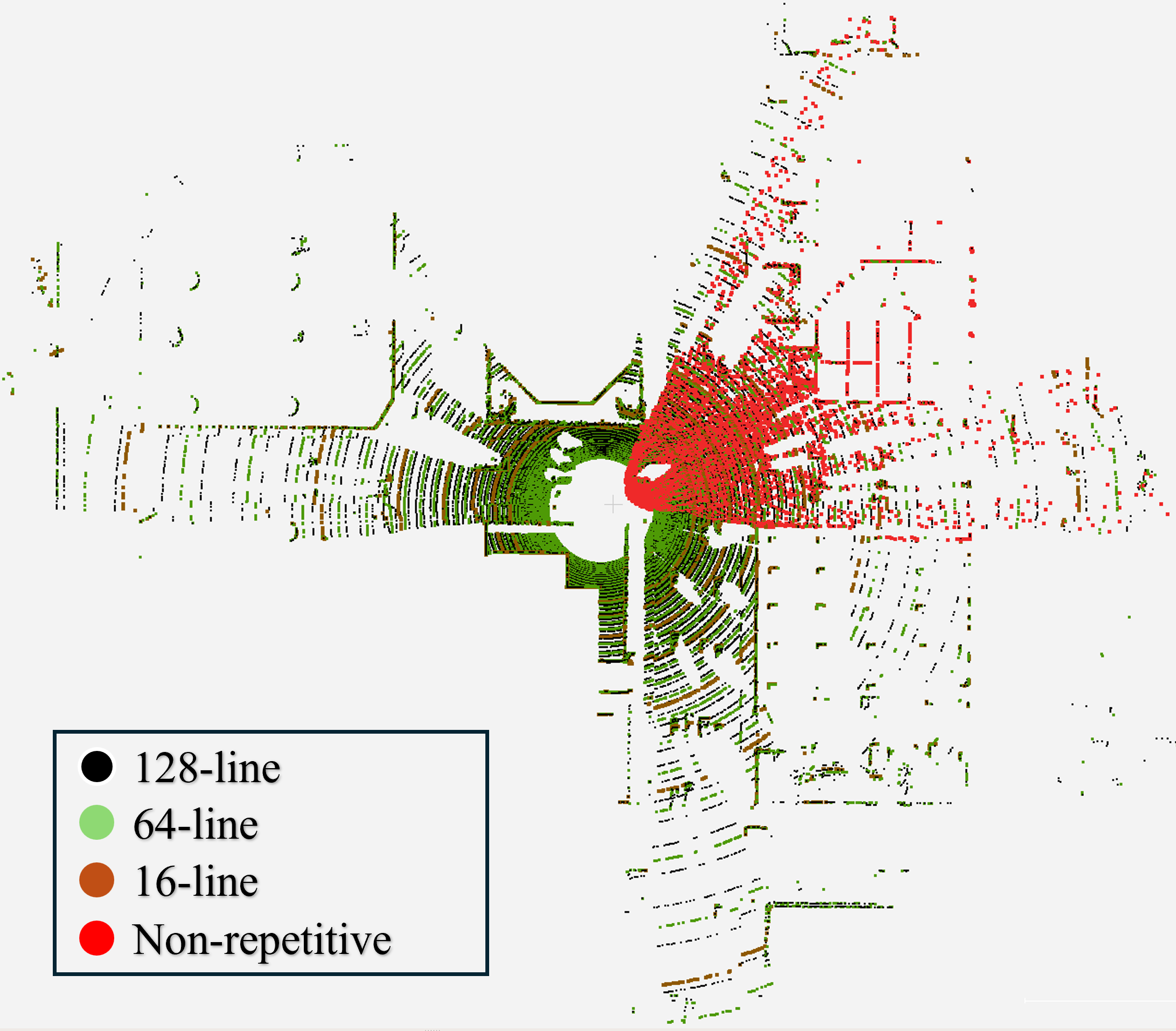} 
        \caption{Crossroad} 
        \label{fig:crossroad_pointcloud} 
    \end{subfigure}
    \hfill 
    \begin{subfigure}[b]{0.2224\textwidth}
        \centering
        \includegraphics[width=\linewidth]{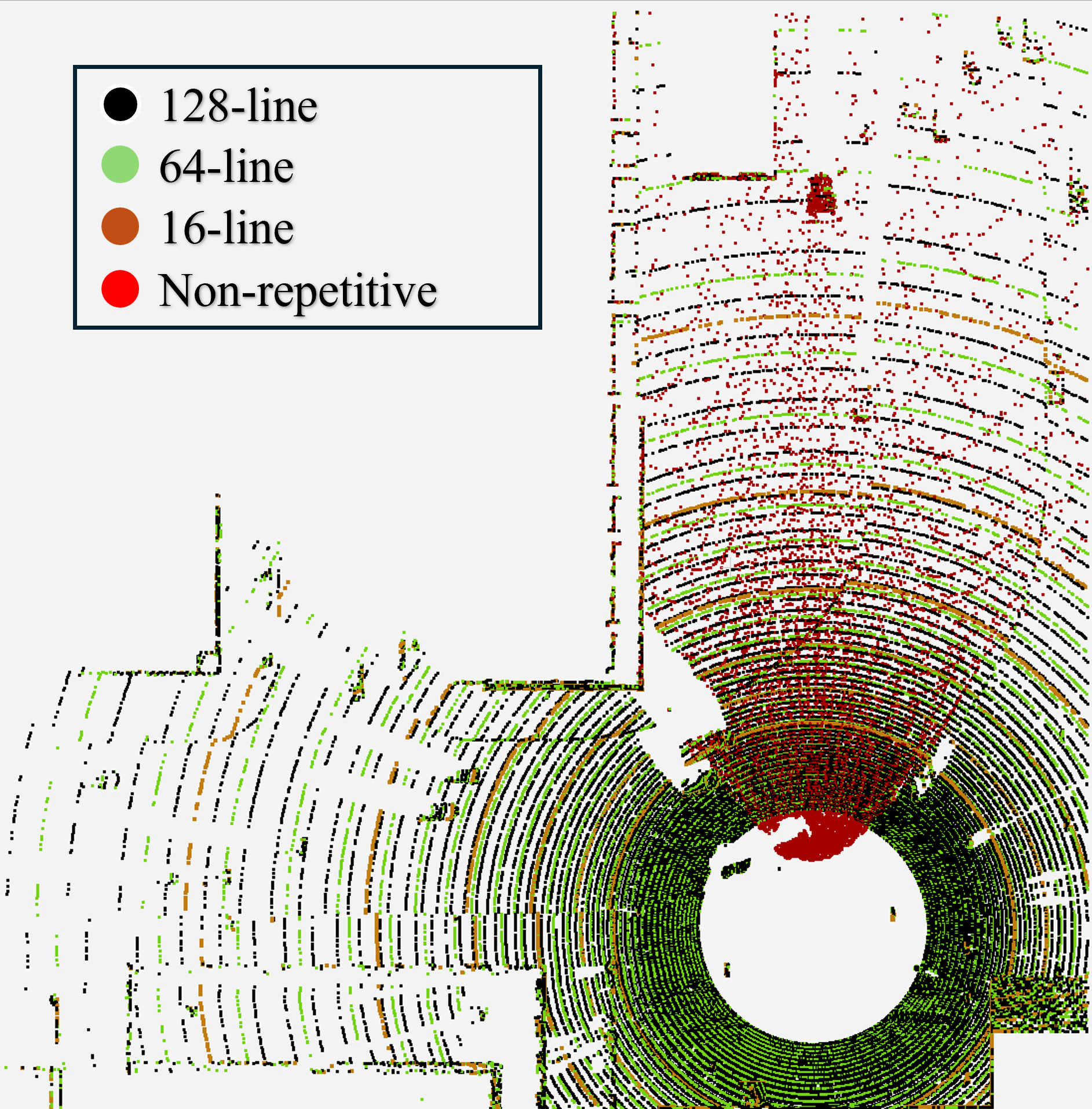} 
        \caption{Curve} 
        \label{fig:curve_pointcloud} 
    \end{subfigure}

    \caption{Visualization of point clouds of different LiDARs} 
    \label{fig:pointcloud_visualization} 
\end{figure}

Our dataset covers three typical scenarios in autonomous driving: highway, crossroad and curve. The three scenarios are shown in Figure \ref{Lidar position}. The highway section was meticulously collected using the official map “Town04,” which provides a realistic highway setting. We randomly placed 300 dynamic vehicles on the map, including various types such as cars, trucks, and taxis. The crossroad data was collected using the official map 'Town05,' while the curve data was collected using the official map 'Town10'. Also we placed 300 dynamic vehicles. Visualization of different LiDARs' point clouds are displayed in Figure \ref{fig:pointcloud_visualization}.

\begin{figure}
    \centering
    \includegraphics[width=1\linewidth]{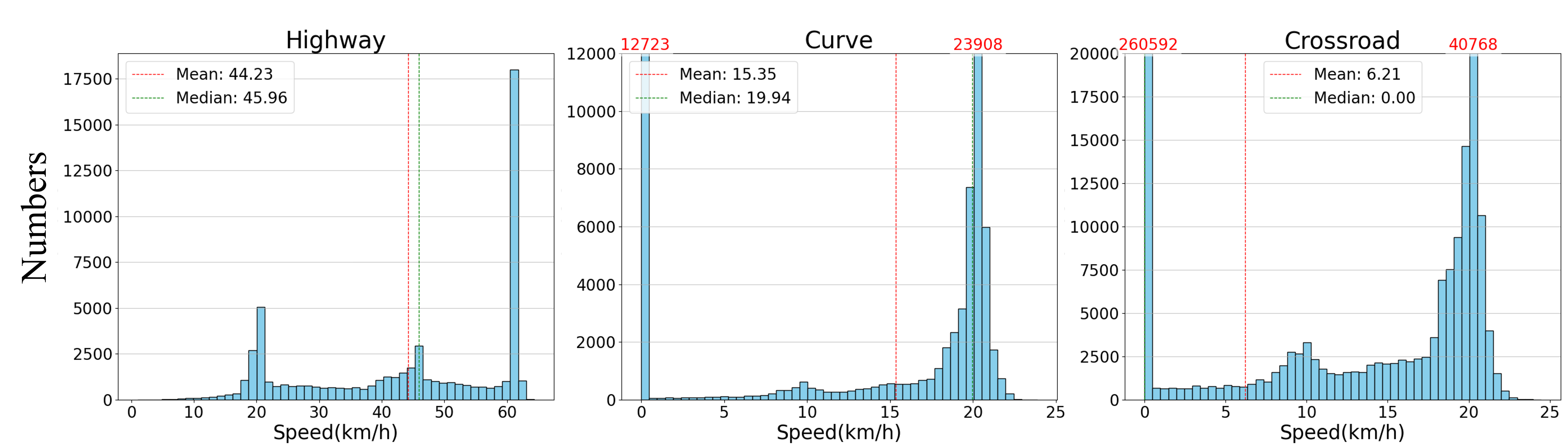}
    \caption{The graph illustrates the velocity distribution of the three scenes. The red numerals above the chart indicate the count of vehicles whose speed exceeded the maximum value displayed on the x-axis.}
    \label{velocity distribution}
\end{figure}

Vehicle velocity is another critical parameter of the simulated scenarios, as it directly influences the quality of the acquired point cloud data. Higher velocities can introduce motion-induced distortions and result in sparser point distributions on moving objects, complicating downstream tasks such as the time deskewing of 3D scans. The velocity distributions for our three experimental scenarios are presented in Figure \ref{velocity distribution}. Due to the traffic management settings within the CARLA environment, vehicle speeds in each scenario are not random but instead tend to cluster around specific velocities. For instance, speeds in the highway scenario are tightly concentrated around 60 km/h, while the crossroad and curve scenarios both exhibit distinct primary speed modes around 20 km/h.

\subsection{LiDAR Simulation Methodology}
Specifically, we concurrently collected data frames using common infrastructure-based LiDARs for our experiments. To ensure fair data comparisons, we positioned the four LiDARs at the same location and at a height of 6m above the ground. Furthermore, the orientation for each LiDAR were standardized: all units were mounted horizontally and aimed to maximize their effective coverage area within the specific scenario. The parameter configurations, such as field of view and points collected per second, were set according to the officially released specifications of commercial models to simulate realistic selection and deployment in real-world roadside scenarios. We have outlined the detailed configurations of the four LiDAR setups and price in Table \ref{lidar_comparison_with_price}. The non-repetitive LiDAR is far cheaper than 64-line and 128-line repetitive LiDARs.

\begin{figure}[htbp] 
    \centering 
    \begin{subfigure}[b]{0.48\textwidth} 
        \centering
        \includegraphics[width=\linewidth]{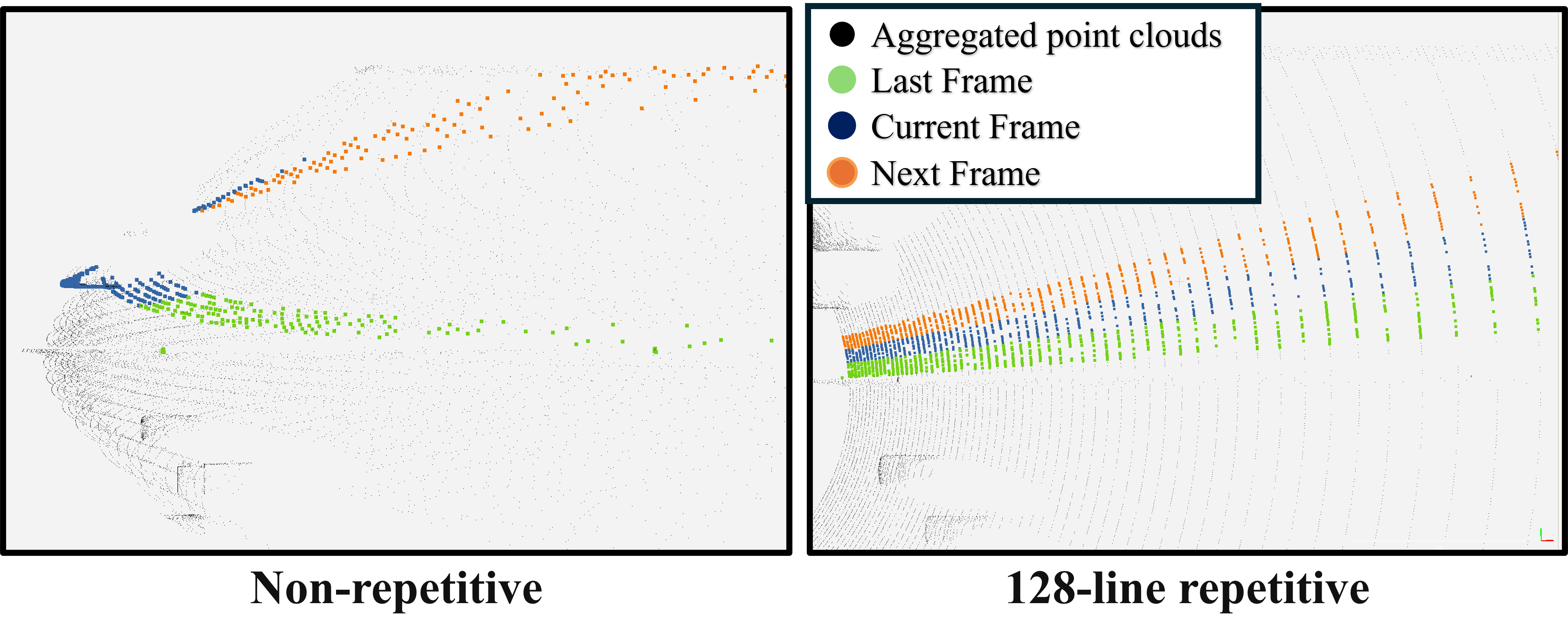} 
        \caption{Aggregated point clouds of different LiDARs} 
        \label{time deskewing} 
    \end{subfigure}
    \hfill
    \begin{subfigure}[b]{0.48\textwidth} 
        \centering
        \includegraphics[width=\linewidth]{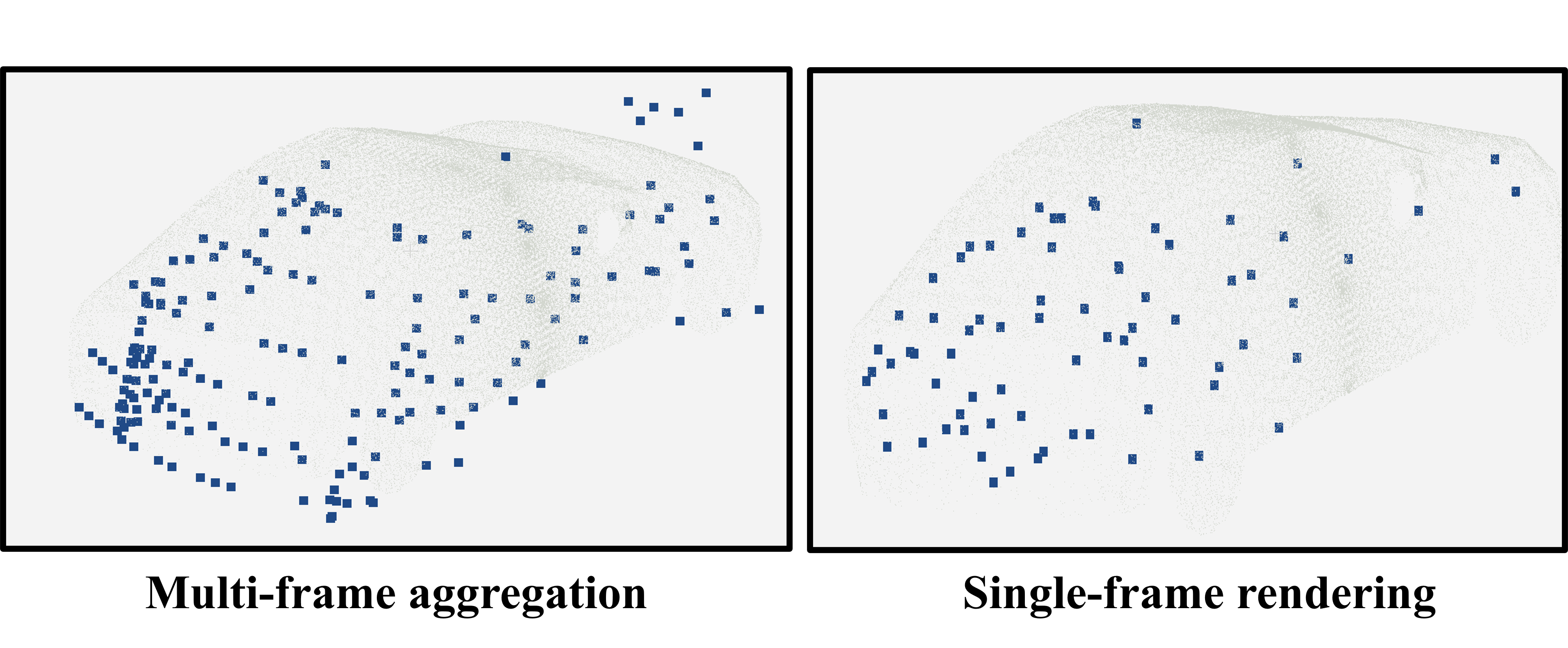} 
        \caption{Difference between two point clouds acquisition methodologies} 
        \label{time deskewing2} 
    \end{subfigure}
    \caption{ (a) illustrates the temporal point cloud aggregation process for a LiDAR sensor, showcasing data from three consecutive frames alongside the denser, cumulative point cloud resulting from 100 frames of aggregation. (b) illustrates how different LiDAR data acquisition methodologies lead to distinct point cloud distributions on an identical target vehicle.} 
    \label{fig:fulltime deskewing} 
\end{figure}

To enhance the fidelity of LiDAR simulation in CARLA, we implemented a point cloud aggregation strategy to mitigate the impact of time deskewing of 3D scans. Conversely, CARLA typically generates point clouds via instantaneous rendering, which can lead to discrepancies from physical sensor behavior. To address this, we configured the simulation time step to 0.001 seconds and aggregated the point clouds generated over 100 consecutive simulation frames as one output frame, which is illustrated in Figure \ref{time deskewing}. This aggregated point cloud serves as the effective output for a single LiDAR scan, thereby more closely emulating real-world sensor data characteristics, particularly the impact of dynamic object motion during the acquisition period. The impact of time deskewing of 3D scans is displayed in Figure \ref{time deskewing2}. Based on this simulation setup, the output frame rate is 10 Hz. We collected 5000 frames for each scenario under the same environmental settings, which means that the data collection period for each scenario was 500 seconds.

Due to their different intrinsic settings, the LiDARs under study generate varying single-frame point counts across the different scenarios, as detailed in Table \ref{avn_lidar_by_scenario}. To ensure a fair and direct comparison, a standardized deployment configuration was adopted for all units. Specifically, all LiDARs were mounted horizontally and given the same forward-facing orientation. This horizontal placement was chosen as a practical baseline; for repetitive scanning LiDARs, it ensures their 360° field of view can be fully utilized for vehicle detection. For non-repetitive LiDARs, minor adjustments to the vertical orientation (e.g., tilting) do not yield a significant overall performance advantage. While tilting the sensor downwards can increase point density in the near field, this over-accumulation offers little benefit for improving detection at farther, more critical distances. Consequently, the choice of vertical angle has a relatively minor impact on the sensor's overall effectiveness, justifying the use of a standardized horizontal deployment.

This standardized placement methodology is also grounded in practicality. In real-world applications, infrastructure-based sensors are typically installed on existing structures, meaning different sensor models would be evaluated from the same position. Therefore, our approach of comparing different LiDARs from an identical location and orientation closely mirrors the real-world challenge of selecting the best sensor for a given deployment site.

\begin{table}[htbp]
    \centering
    \begin{threeparttable}
        \caption{Setups and Estimated Prices for LiDARs Used in Dataset Collection}
        \label{lidar_comparison_with_price} 
        
        \begin{tabular}{@{}llcll@{}} 
            \toprule
            LiDAR Name & LiDAR Type & FOV $H \times V$ & Points/s & Price\tnote{e} \\
            \midrule
            Livox Avia\tnote{a} & non-repetitive      & $77.2^{\circ} \times 70.4^{\circ}$  & 240,000   & \$2,079  \\
            Helios 16\tnote{b}  & 16-line repetitive  & $360^{\circ} \times 30^{\circ}$     & 288,000   & \$2,399  \\
            Pandar 64\tnote{c}  & 64-line repetitive  & $360^{\circ} \times 40^{\circ}$     & 1,152,000 & \$6,425 \\
            Ruby Plus\tnote{d}  & 128-line repetitive & $360^{\circ} \times 40^{\circ}$     & 2,304,000 & \$24,000 \\
            \bottomrule
        \end{tabular}
        \begin{tablenotes}
            \item[a] \footnotesize{\url{https://www.livoxtech.com/cn/avia}}
            \item[b] \footnotesize{\url{https://www.robosense.cn/IncrementalComponents/Helios}}
            \item[c] \footnotesize{\url{https://www.hesaitech.com/product_downloads/pandar64-4/}}
            \item[d] \footnotesize{\url{https://www.robosense.cn/rslidar/RubyPlus}}
            \item[e] \footnotesize{Prices are estimates based on publicly available data and are for reference only. Actual prices may vary significantly based on vendor and volume.}
        \end{tablenotes}
    \end{threeparttable}
\end{table}

\begin{table}[htbp]
    \centering
    \begin{threeparttable}
        \caption{Average Number of Hit Points per Scan} 
        \label{avn_lidar_by_scenario} 
        \begin{tabular}{@{} L{2.5cm} R{1.5cm} R{1.5cm} R{1.5cm} @{}} 
            \toprule
            \textbf{LiDAR Type}         & \textbf{Highway} & \textbf{Crossroad} & \textbf{Curve} \\
            \midrule
            Non-repetitive        & 10,529           & 11,120             & 10,953         \\
            16-line Repetitive    & 2,908            & 6,153              & 5,081          \\
            64-line repetitive    & 23,344           & 33,873             & 30,585         \\
            128-line repetitive   & 46,723           & 68,271             & 61,399         \\
            \bottomrule
        \end{tabular}
    \end{threeparttable}
\end{table}

\section{Statistical Benchmark based on LiDAR Scanning Pattern} 
In the first stage of our experiments, we conducted a comprehensive comparison of the scanning capabilities across four selected LiDAR systems. Our investigation centered on quantifying the number of points each LiDAR captured when detecting vehicles. To systematically evaluate object capture quality, we established classification thresholds at 5 and 20 points: fewer than 5 points indicated low-quality detection, 5 to 20 points represented medium-quality detection, and more than 20 points signified high-quality detection.

To further quantify perception performance, we introduce the "Average Points per Vehicle" (APV) metric, which measures a LiDAR's mean data acquisition capability on targets within specific distance ranges (e.g., 60-80 meters). This metric is calculated as the ratio of the total number of points captured on vehicles (NP) to the total number of vehicles (NV) within that same range, formulated as $APV=\frac{NP}{NV}$.

To enhance classification accuracy, we implemented two preprocessing steps: first, we filtered out all points below the horizon ($z \leq 0$), and second, we refined certain vehicle bounding boxes in CARLA that failed to fully enclose the vehicles. The resulting detection metrics—both quantitative (point counts) and qualitative (detection quality classifications)—formed the basis for our comparative analysis in subsequent experiments.

We selected the highway scenario in CARLA as our testing environment due to its unobstructed sightlines for the LiDARs and the linear road geometry, which facilitated more straightforward distance-based analysis from the sensor position.

\subsection{Qualitative Analysis on LiDAR Scanning Ability}

\begin{figure}[htbp]
    \centering
    \includegraphics[width=1\linewidth]{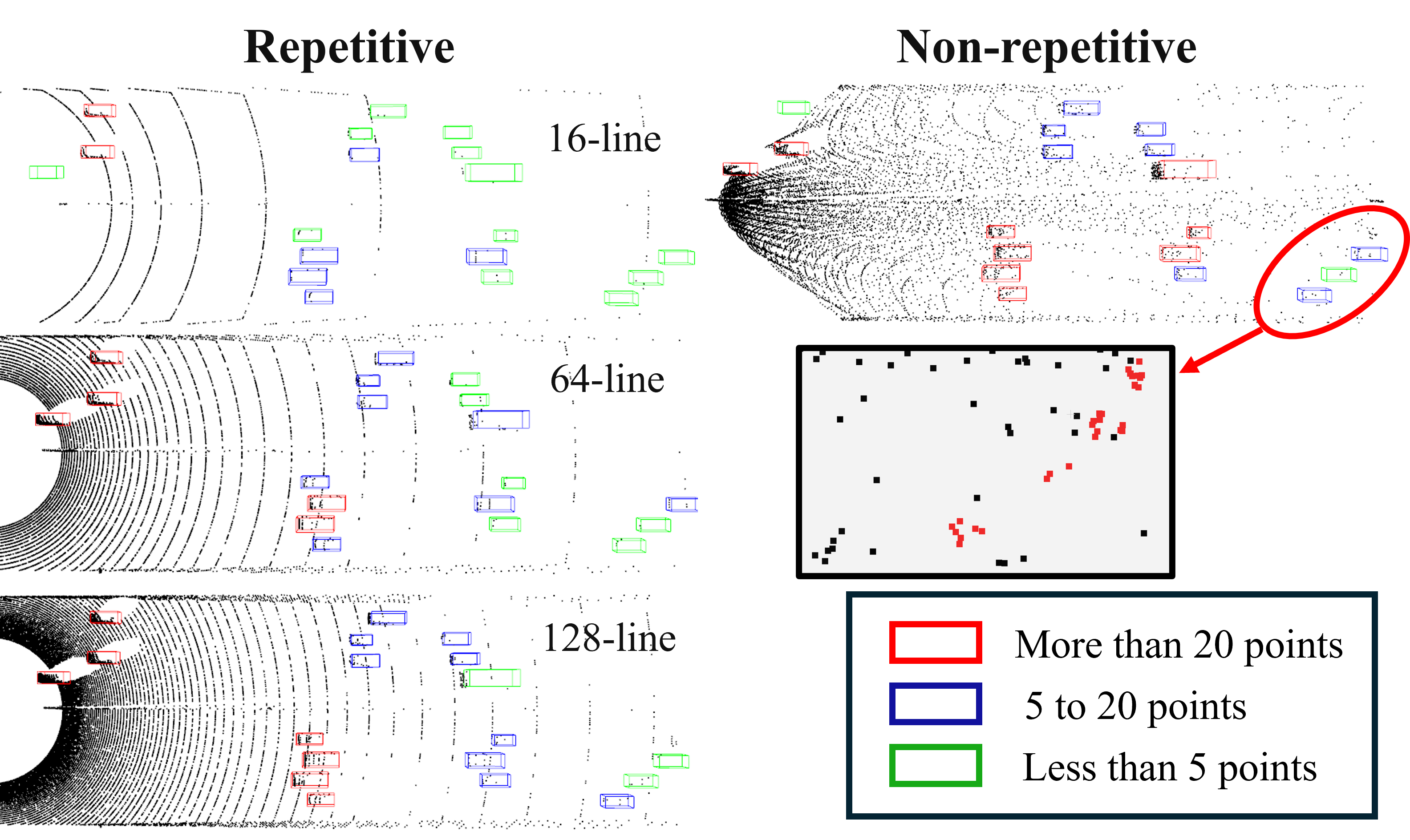}
    \caption{Visualization of detected objects, indicated by color-coded bounding boxes. The arrow indicates an enlarged view of the circled area. In the enlarged view, the red dots represent the points inside the detection box,while the white dots represent the points outside the detection box.}
    \label{Qualitative Analysis on LiDAR Scanning Ability}
\end{figure}

Figure \ref{Qualitative Analysis on LiDAR Scanning Ability} presents a comparative visualization of the four LiDARs object detection capabilities. The figure displays captured object bounding boxes color-coded according to point density: green boxes indicate low-density detections (fewer than 5 points), blue boxes represent medium-density detections (5 to 20 points), and red boxes signify high-density detections (more than 20 points).

The regions highlighted with red circles reveal a significant finding: the non-repetitive LiDAR successfully detects vehicles at extended ranges where all repetitive-scanning LiDAR systems fail to register any detection. This enhanced long-range detection capability can be attributed to the distinctive non-repetitive scanning pattern, which provides superior detection performance at greater distances.

When comparing the three repetitive-scanning LiDARs: an anticipated quantitative trend emerges: the average number of points captured per detected object sequentially decreases with the reduction in line count. This decline in average point density across the three LiDARs also follows a consistent pattern, a characteristic linked to their fundamentally similar repetitive scanning mechanisms.

\subsection{Quantitative Analysis on LiDAR Scanning Ability}

\begin{figure}[htbp]
    \centering
    \includegraphics[width=1\linewidth]{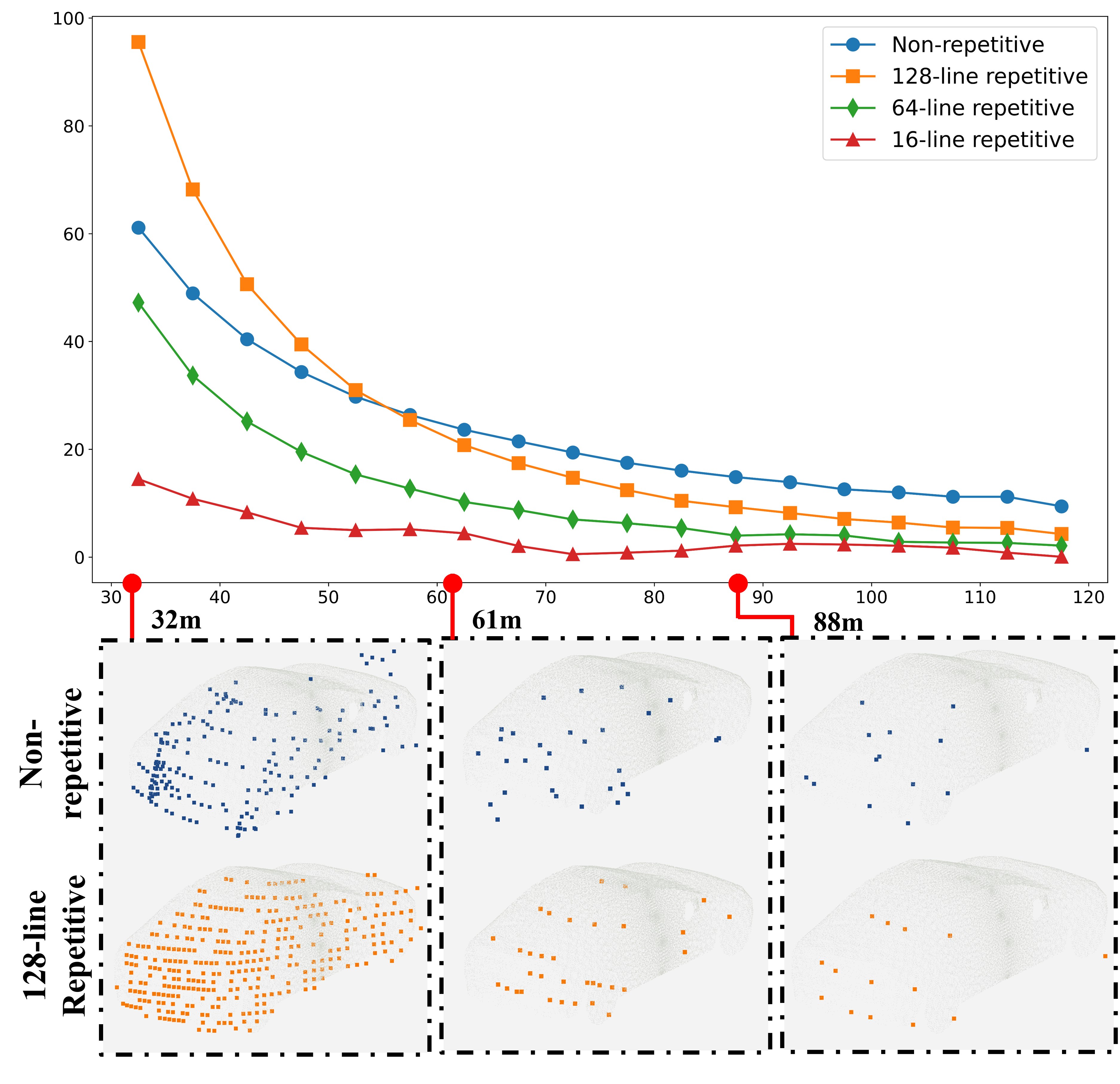}
    \caption{Average point cloud numbers at different distances to the LiDAR. The pointed-out parts are visualization of the vehicle-captured point clouds.}
    \label{average_number}
\end{figure}

We excluded points within a 30-meter radius of the LiDAR sensor due to field-of-view blind spots. The remaining point cloud data was categorized into 5-meter intervals ranging from 30 to 120 meters. Figure \ref{average_number} illustrates the average number of vehicle point clouds captured within each interval. The results clearly demonstrate that the non-repetitive LiDAR outperforms the other three repetitive LiDAR systems in detecting distant vehicles. At ranges exceeding 90 meters, the non-repetitive LiDAR captures approximately twice the number of points compared to the alternative repetitive LiDAR systems. Thus, in this section, we can draw the following conclusions:

\begin{itemize}
    \item 
    When positioned closer to the LiDAR, the 128-line repetitive LiDAR demonstrates superior performance.

    \item 
    Between the three repetitive-scanning LiDARs, the 128-line, 64-line, 16-line LiDARs maintain an approximate 8:4:1 ratio in average point cloud numbers, which aligns with their 'Points/s' configurations.
    
    \item 
    The non-repetitive LiDAR, exhibits significantly better perception capabilities at longer distances compared to the other three systems, with a notably gentler degradation curve as distance increases.

\end{itemize}

\section{Performance Benchmark based on Perception Algorithms}

In the second stage of our experiments, we utilized objects that were successfully captured by LiDAR as ground truth to train and test multiple 3D object detection algorithms. We aimed to investigate the impact of various point cloud patterns generated from different infrastructure LiDARs on model performance, across different data representation paradigms and backbone architectures.

While the preceding analyses focused on the intrinsic data acquisition characteristics and static scanning capabilities of the different LiDAR systems, their ultimate efficacy in real-world roadside deployments is determined by their ability to support robust dynamic perception tasks. To this end, in the second stage of our experiments, we utilized objects that were successfully captured by LiDAR as ground truth to train and test multiple 3D object detection algorithms. We aimed to investigate the impact of various point cloud patterns generated from different infrastructure LiDARs on model performance, across different data representation paradigms and backbone architectures.

\subsection{Algorithm Configurations}

We selected specific algorithms for each paradigm: PointRCNN for point-based algorithms, PointPillars for pillar-based models, PV-RCNN for voxel-based models, and DSVT for innovative transformer-based two-stage detectors. The features of different algorithms are shown in Table \ref{tab:algorithm_features_comparison}. As for the training configurations of our experiments, the configuration of the algorithm remains the same for different LiDARs. After comparing the model performance with the coverage range, we figure out under certain circumstances, what's the performance of each LiDAR.

\begin{table}[htbp]
    \centering
    \caption{ Features of Selected 3D Object Detection Algorithms}
    \label{tab:algorithm_features_comparison}
    \setlength{\tabcolsep}{4pt} 
    \begin{tabular}{@{} l L{2.5cm} L{4.5cm} @{}} 
        \toprule
        Algorithm    & Architecture Type / Stages        & Key Mechanisms                                        \\
        \midrule
        PointRCNN    & Point-based / Two-stage           & PointNet++ for proposal generation, canonical coordinate refinement  \\
        \addlinespace
        PointPillars & Pillar-based / Single-stage & PointNet-like operations on pillar points, 2D CNN on BEV map  \\
        \addlinespace
        PV-RCNN      & Voxel-based / Two-stage & 3D Sparse CNN backbone, Voxel Set Abstraction module combining voxel and point features  \\
        \addlinespace
        DSVT         & Transformer-based / Single-stage & Fully Sparse Voxel Transformer backbone, dynamic attention mechanisms \\
        \bottomrule
    \end{tabular}
\end{table}

\subsection{Evaluations}

To quantitatively assess the performance of various algorithms, our evaluation methodology is predicated on Average Precision (AP) as the primary metric. As for the definition of AP, we consider the vehicles identified by the algorithm models as true positive (TP) vehicles. In particular, we have defined predicted true positive instances as those having prediction boxes with an intersection over union (IoU) value of 0.5 or greater when compared to the corresponding ground truth boxes. Considering that the data is directly detected by the LiDAR sensors in the field, and vehicle localization does not heavily rely on the vertical positioning of vehicles, we chose the Bird's Eye View (BEV) perspective for the evaluation, where IoU calculations are based on the 2D projections of vehicles onto the ground plane, providing a top-down view that aids in better assessing vehicle position and dimensions. During the evaluation process, we exclude the boxes with scores that fall below 0.1, which helps filter out low-confidence predictions and enhances the reliability of the assessment. It is essential to note that the 3D object detection algorithms detect only a subset of the vehicles captured by the LiDAR. It is important to note that the ground truth boxes differ between the non-repetitive LiDAR and the other three systems. This is because the non-repetitive LiDAR has a limited Horizontal Field of View (HFOV) covering the region directly in front of it, whereas the other three LiDARs have a $360^\circ$ HFOV. 

The average precision(AP) is defined as:
$$AP=\frac{1}{40}\sum_{r\in R}\frac{TP}{TP + FP}$$
where we use the “$AP|_{R_{40}}$" to calculate the "$AP$". The “R” represents a recall set containing 40 sequential recall
 points,and the “FP” represents the number of false positive estimated vehicles. Based on the definition of AP, we have devised three distinct analytical approaches to provide a multifaceted examination of algorithmic efficacy: Overall AP Analysis, Distance-Segmented AP Analysis, and High-Quality Detection Area analysis. It is worth noting that the models were trained and tested on their respective training and testing sets with the same LiDAR configuration setup. For instance, we evaluated the results of the “PointPillars” model on the “non-repetitive” testing frames, having been trained on the “non-repetitive” training frames.

\subsubsection{Overall AP Analysis}
The Overall AP Analysis provides a quantitative measure of detection accuracy by directly comparing predicted vehicle bounding boxes with their ground truth counterparts within the LiDAR's operational field of view. In this metric, higher values are unequivocally better, indicating the algorithm's enhanced capability to accurately delineate target objects.

\subsubsection{Distance-Segmented AP Analysis}
Referring to the previous comparative experiments on point cloud density at varying distances, we noted disparities in distance-based attenuation between the non-repetitive LiDAR and other repetitive scanning LiDAR point clouds. To investigate whether these differences persist under various algorithm prediction conditions, we conducted further research under the following experimental parameters.

\begin{figure}[htbp]
    \centering
    \includegraphics[width=1\linewidth]{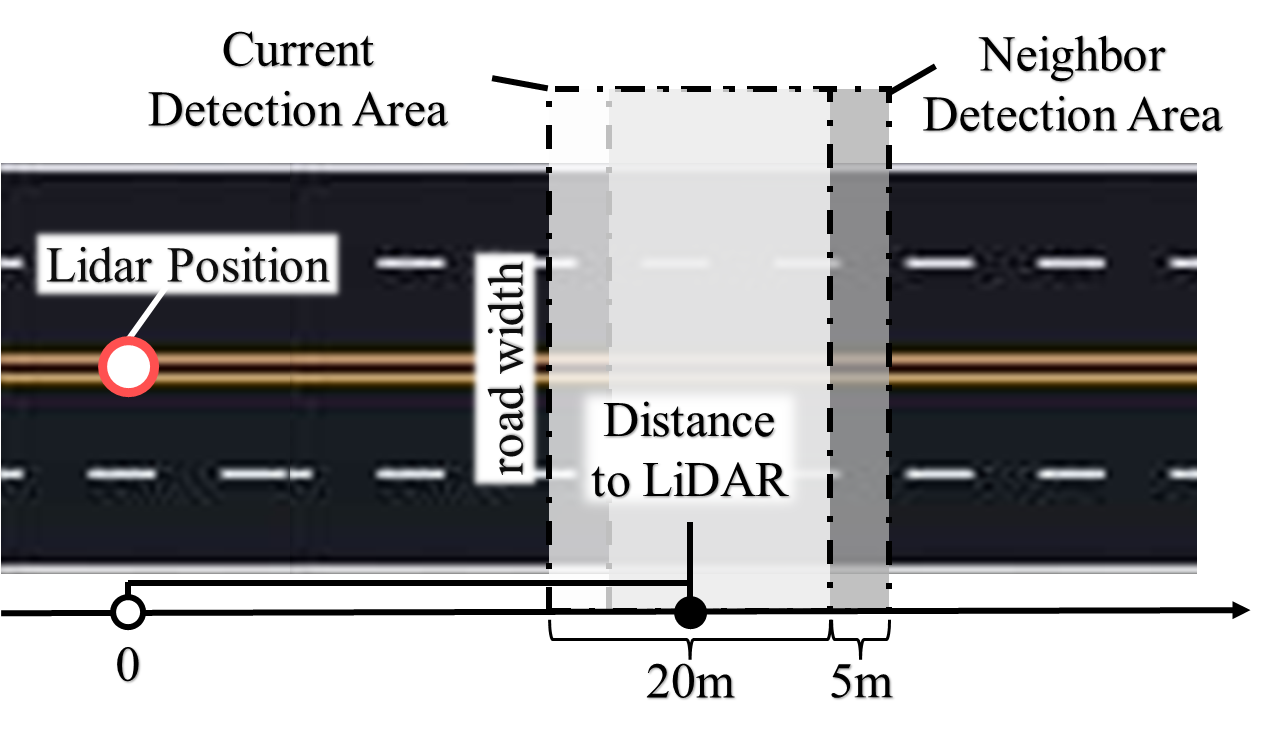}
    \caption{LiDAR Detection Area Illustration}
    \label{ROI showcast}
\end{figure}

The Distance-Segmented AP Analysis is only conducted in the highway scenario because of its relatively straight road configuration in highway scenarios. It is conducive to directly analyze the performance of different algorithms on vehicle point clouds at varying distances from the LiDAR. We implemented a sampling strategy at 5-meter intervals, calculating the Average Precision (AP) for vehicles within a 20-meter area along the lane direction, centered on each sampling point. This approach is illustrated in Figure \ref{ROI showcast}. The AP value at each sampling point is representative of the algorithm's performance within its corresponding 20-meter area. For example, the AP attributed to a sampling point at 40 meters represents the algorithm's detection efficacy for vehicles located in the 20-meter zone between 30 and 50 meters from the LiDAR installation.

\subsubsection{High-Quality Detection Area Analysis}

Given the varying horizontal field of view (HFOV) among different LiDAR systems, calculating Average Precision (AP) alone does not provide a comprehensive evaluation of LiDAR performance. The non-repetitive LiDAR has a notably limited horizontal HFOV of 70.4°, which contrasts significantly with the 360° coverage of rotational scanning LiDARs. This discrepancy manifests particularly in highway scenarios, where non-repetitive LiDAR can only detect vehicles directly in front of it, while the other three LiDAR systems are capable of detecting vehicles across the entire roadway.

Based on these considerations, we quantified the number and proportion of regions with an Average Precision (AP) exceeding 0.85, which we define as high-quality detection area. For the selection of regions, we adopt the same segmentation strategy used in the Distance-Segmented AP Analysis. Specifically, the roadway is evaluated using a series of overlapping areas, each measuring 20 meters in length and spanning the width of the road. These areas are sampled at 5-meter intervals along the direction of travel. To provide a concrete example: for a 200-meter long highway scene, this methodology yields a total of $200 \div 5=40 $ detection areas. Each of these 40 areas is 20 meters long and is centered on its respective sampling point.

We will evaluate the following key quantities: Total Number of Detection Areas ($N_{TDA}$), which represents the total count of discrete segments defined by our evaluation methodology for a given scene, Number of High-Quality Detection Areas ($N_{HQDA}$), which refers to the count of areas where the detection performance, measured by Average Precision (AP), is greater than 0.85. The High-Quality Detection Area Proportion ($P_{HQDA}$) is the ratio of the Number of High-Quality Detection Areas to the Total Number of Detection Areas. Their relationship is formulated by the following Equation \ref{PHQDA}.

\begin{equation}
\label{PHQDA}
P_{HQDA} = \frac{N_{HQDA}}{N_{TDA}}
\end{equation}

\subsection{Standardized Experiments in the Highway Scenario}
The highway scenario was selected for our standard performance analysis primarily due to its relatively straight road configuration. This characteristic is particularly conducive to unambiguous distance-dependent performance evaluations, a methodology central to our Distance-Segmented AP Analysis. In our experiment on the highway scenario, the dataset was divided into 3348 training frames and 838 testing frames. We excluded 652 frames from the algorithm testing stage because they did not have any vehicles within the specified range. The Overall Average Precision (Overall AP) results obtained for this scenario are presented in Table \ref{highway_AP}.

\begin{table}[htbp]
    \centering
    \begin{threeparttable}
        \caption{The performances of different 3D object detection algorithms in the highway scenario.}
        \label{highway_AP}
        \begin{tabular}{lcccc} 
            \toprule
            LiDAR Type & PointRCNN & PointPillars  & PV-RCNN & DSVT  \\
            \midrule
            16-line repetitive& 26.16 & 55.02 & 57.56 & 52.54 \\
            64-line repetitive& 36.73 & 83.93  & 87.40 & 80.05 \\
            128-line repetitive& 37.41 & \textbf{92.43}  & \textbf{93.29} & \textbf{83.57} \\ 
            Non-repetitive & \textbf{40.18} & 91.37 & 92.58 & 71.93 \\
            \bottomrule
        \end{tabular}
    \end{threeparttable}
\end{table}

\begin{figure}[htbp]
    \centering
    \includegraphics[width=1\linewidth]{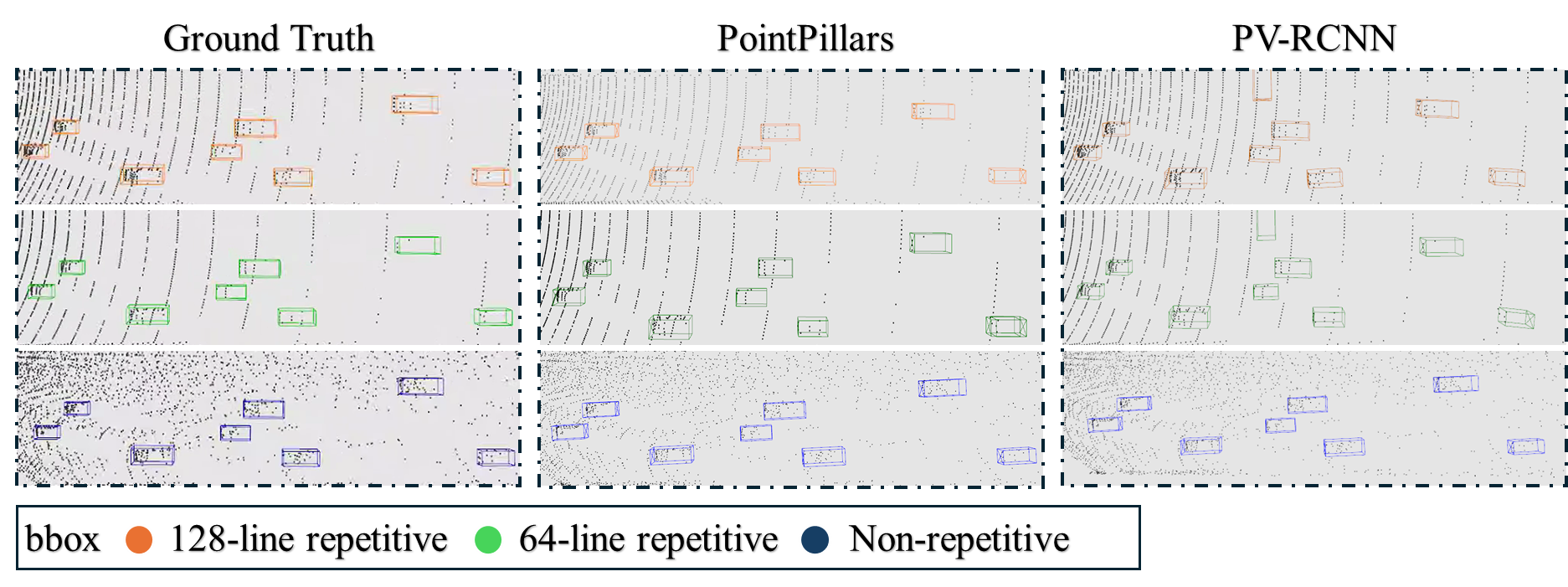}
    \caption{Visualization of Highway Predicted Bbox}
    \label{highway compare}
\end{figure}

As is shown in Table \ref{highway_AP}, we observed that two methods, PV-RCNN and PointPillars, demonstrated superior performance across various LiDAR systems. Consequently, Figure \ref{highway compare} illustrates the bounding boxes predicted by PV-RCNN and PointPillars methods in the highway scenario.

\begin{figure}[htbp]
    \centering
    \includegraphics[width=1\linewidth]{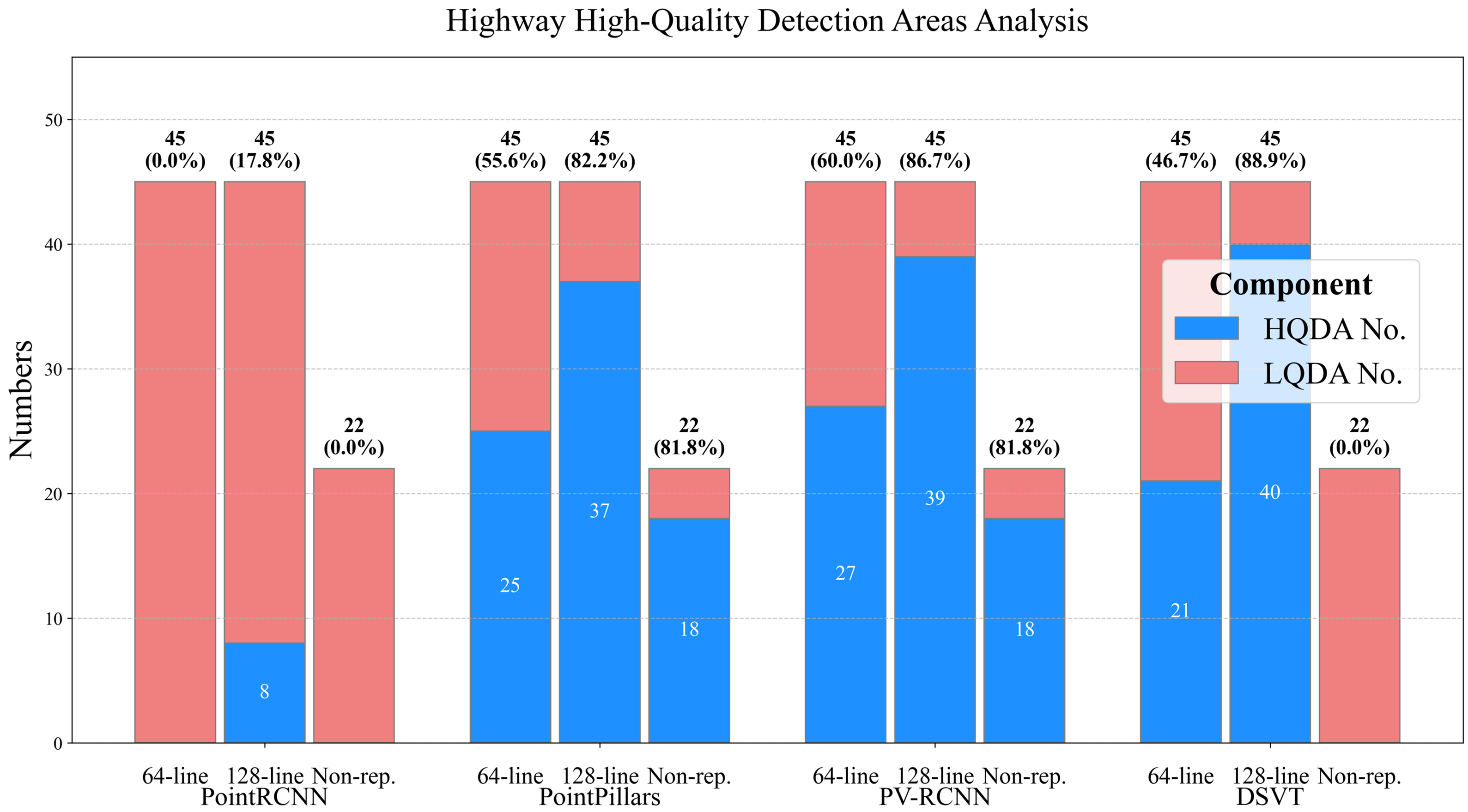}
    \caption{Highway High-Quality Detection Areas Analysis. }
    \label{Highway HQDA}
\end{figure}

The result of the Distance-Segmented AP Analysis is shown in Figure \ref{highway_distance_compare}. And the High-Quality Detection Area analysis is shown in Figure \ref{Highway HQDA}. In this section, we can draw the following conclusions:
\begin{figure}[htbp]
    \centering
    \includegraphics[width=1\linewidth]{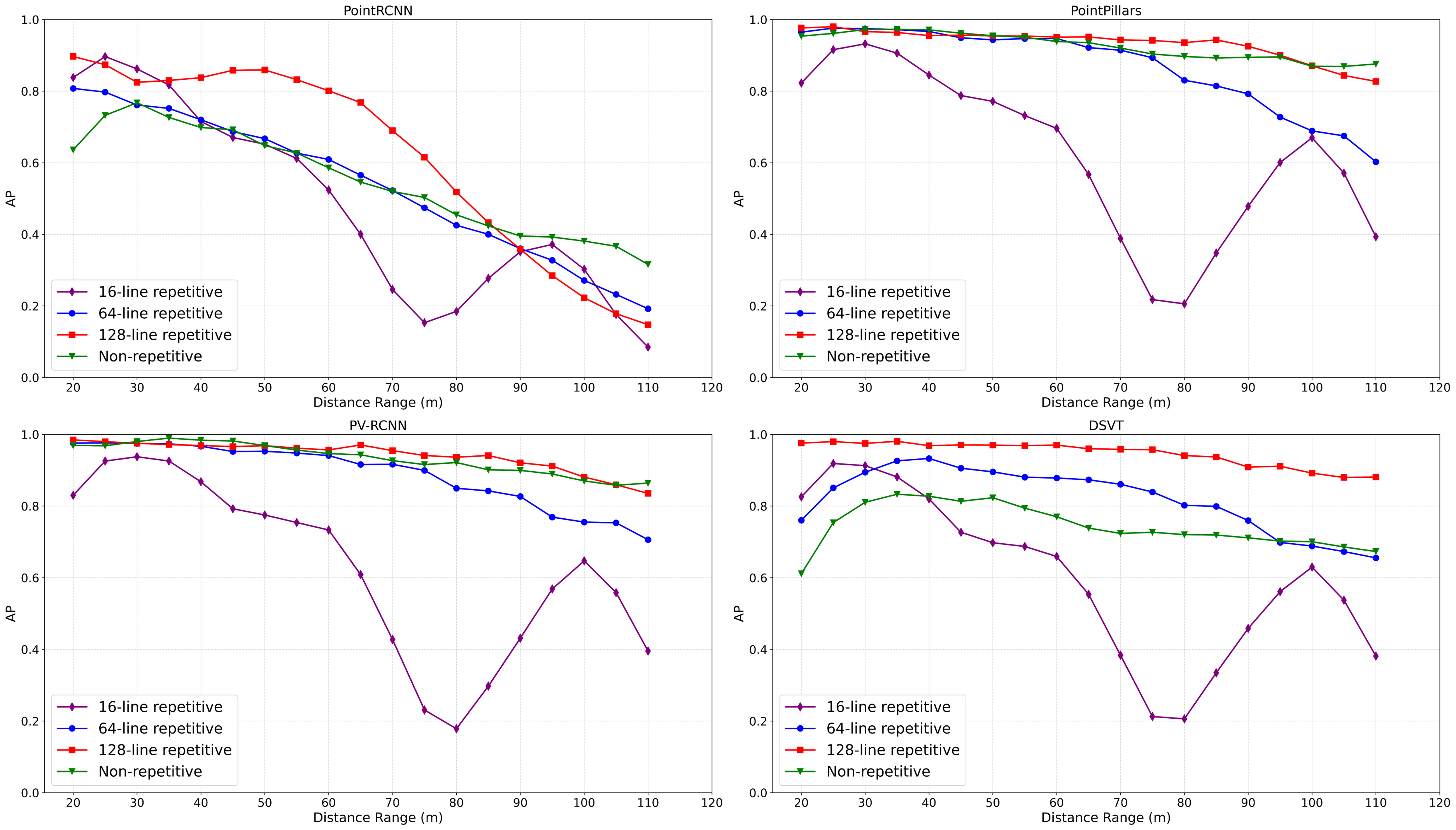}
    \caption{Distance-Segmented AP Analysis in the Highway scenario}
    \label{highway_distance_compare}
\end{figure}

\begin{itemize}
    \item 
    Across most evaluated algorithms, the performance of the non-repetitive LiDAR was broadly comparable to that of the 128-line repetitive LiDAR. In terms of performance, they were followed by the 64-line and 16-line repetitive LiDARs, respectively. 128-line repetitive and non-repetitive LiDARs have a similarly high proportion of high-AP districts. However, due to the HFOV restriction, 128-line repetitive has the highest numbers of high AP district, followed by 64-line repetitive and non-repetitive.
    
    \item All systems exhibit a trend of decreasing Average Precision (AP) with increasing distance. Non-repetitive LiDAR particularly excels at greater distances due to its distinctive point cloud characteristics - a significantly lower point density attenuation rate with increasing distance, which enables it to maintain a greater number of high-AP regions at farther distances from the sensor, correspondingly translating to a higher proportion of such High-Quality Detection Areas. 
    
    \item 
    Considering the different algorithms, PointRCNN performs the worst, while PV-RCNN and PointPillars have a similar better performance. DSVT has a slightly worse performance than the above two algorithms.

\end{itemize}

\subsection{Experiments in the Crossroad and Curve scenarios }
\begin{figure}[htbp]
    \centering 

    \begin{subfigure}[b]{0.48\textwidth}
        \centering
        \includegraphics[width=\linewidth]{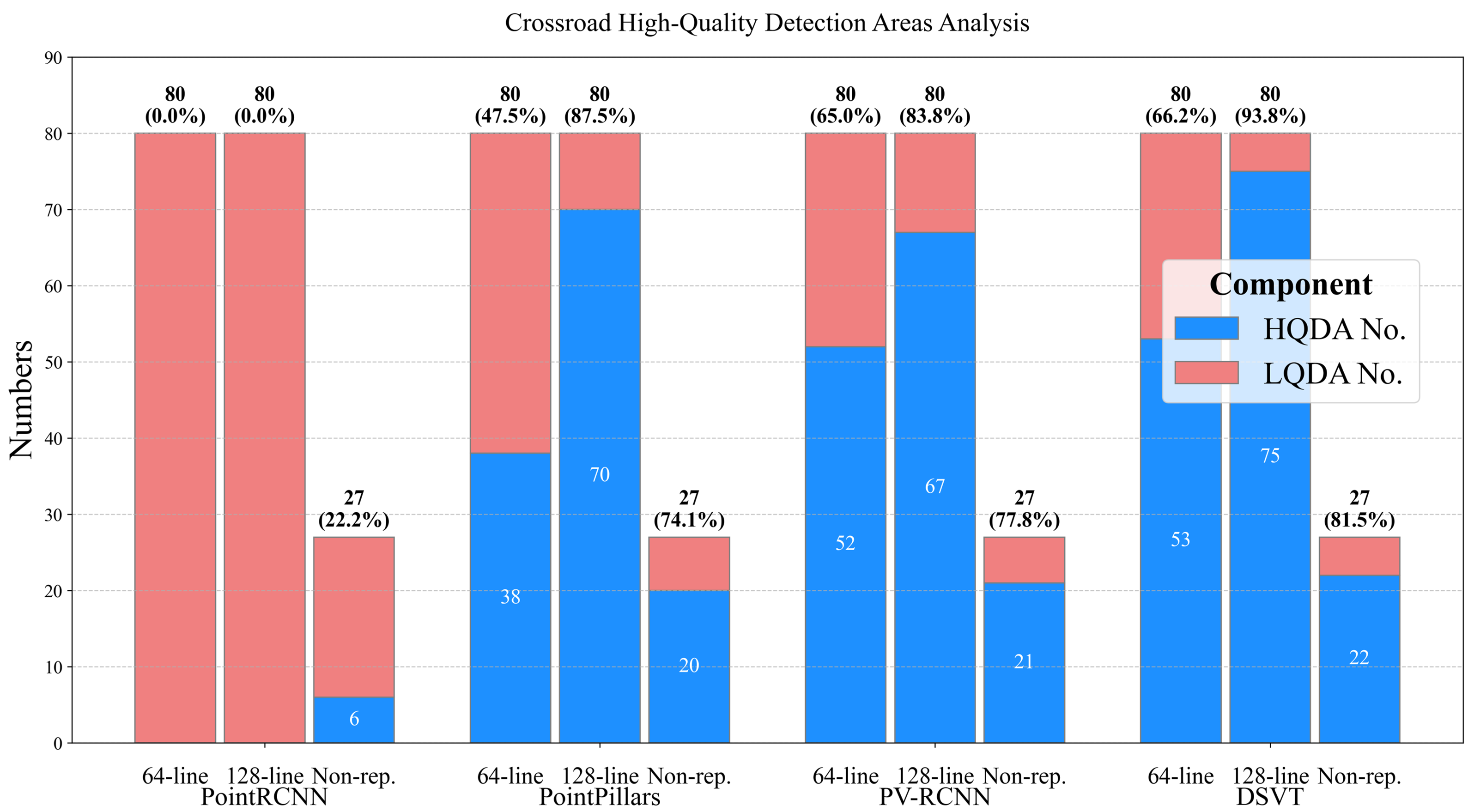} 
        \caption{Crossroad}
        \label{fig:first_sub_figure}
    \end{subfigure}
    \hfill 

    \begin{subfigure}[b]{0.48\textwidth}
        \centering
        \includegraphics[width=\linewidth]{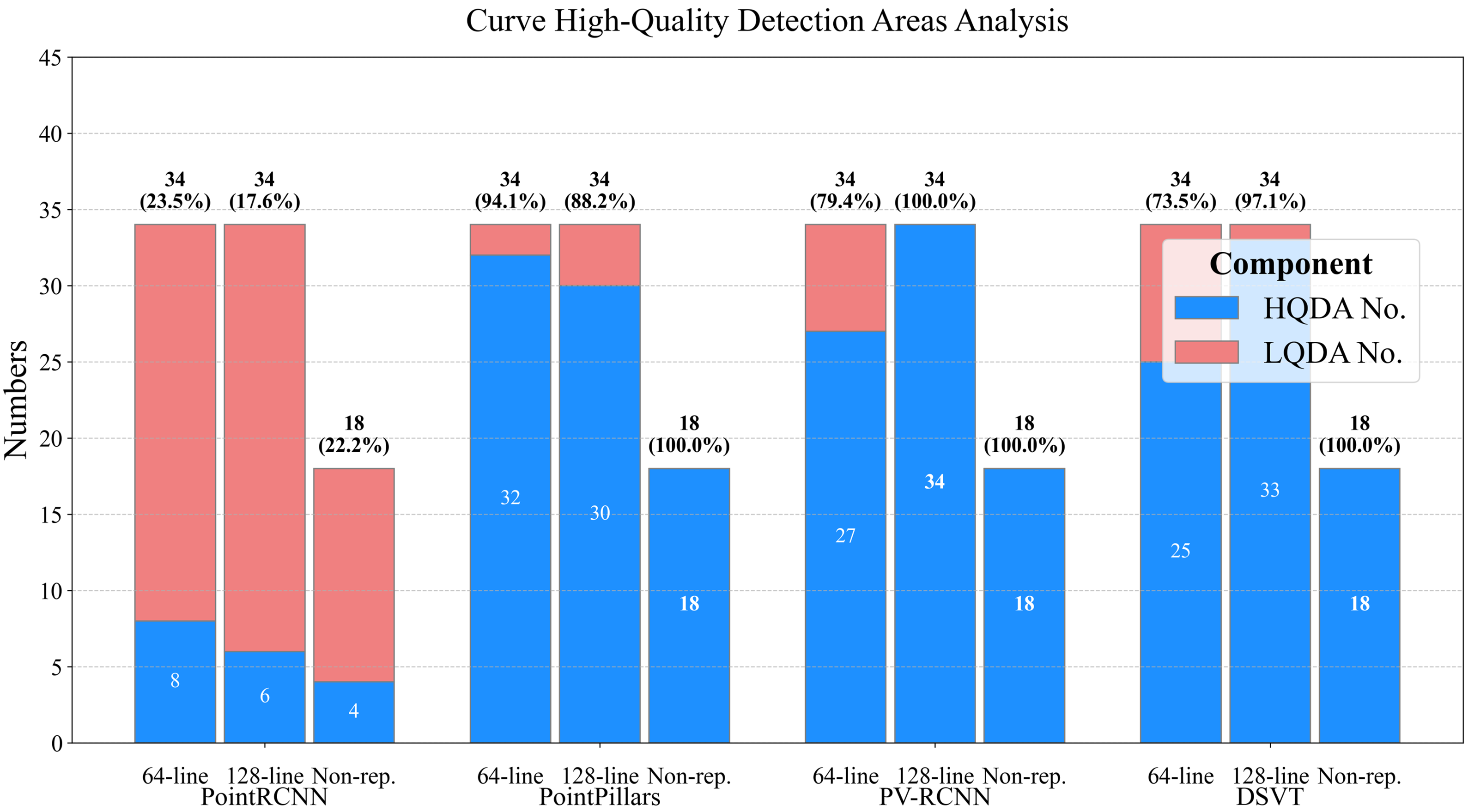} 
        \caption{Curve}
        \label{fig:second_sub_figure}
    \end{subfigure}

    \caption{Analysis of High-Quality Detection Areas. (a) shows that in the crossroad scenario, while (b) shows that in the curve scenario.}
    \label{curve_and_crossroad_HQDA}
    
    \begin{minipage}{1\linewidth} 
        \vspace{3pt} 
        \footnotesize 
        \setlength{\parindent}{0pt} 

        \textbf{a)} HQDA No.: Number of High-Quality Detection Areas. \\
        \textbf{b)} LQDA No.: Number of Low-Quality Detection Areas. \\
        \textbf{c)} TDA No.: Number of Total Detection Areas. TDA = HQDA + LQDA.
    \end{minipage}
\end{figure}

For the crossroad and curve scenarios, we conducted a focused set of experiments, specifically performing Overall AP Analysis and High-Quality Detection Area analysis. In the experiment on both crossroad and curve scenarios, the datasets were divided into 4000 training frames and 1000 testing frames. All of the frames have vehicles within the specified range. The respective results from these evaluations are presented in Table \ref{tab:crossroad_curve_AP} and Figure \ref{curve_and_crossroad_HQDA}.

\begin{table}[htbp]
    \centering
    \begin{threeparttable}
        \caption{Performance of 3D Object Detection Algorithms in Crossroad and Curve Scenarios.}
        \label{tab:crossroad_curve_AP}
        
        \begin{tabular}{@{} L{1.8cm} L{1cm} C{1.05cm} C{1cm} C{1.23cm} L{0.75cm} @{}} 
            \toprule
            LiDAR Type & Scenario & PointRCNN & PointPillars & PV-RCNN & DSVT  \\
            \midrule
            \multirow{2}{*}{16-line repetitive} 
                                 & Crossroad & 9.34 & 31.49 & 39.63 & 30.28 \\
                                 & Curve     & 16.40 & 57.66 & 64.37 & 64.07 \\
            \addlinespace 
            \multirow{2}{*}{64-line repetitive}
                                 & Crossroad & 12.18 & 60.05  & 71.01 & $74.11$\\
                                 & Curve     & 37.23 & 82.29  & 92.84 & 85.35 \\
            \addlinespace
            \multirow{2}{*}{128-line repetitive}
                                 & Crossroad & 13.44 & 78.49  & 77.27 & $79.27$ \\
                                 & Curve     & \textbf{37.70} & 86.99  & \textbf{95.44} & 91.85 \\
            \addlinespace
            \multirow{2}{*}{Non-repetitive}
                                 & Crossroad & \textbf{37.00} & \textbf{84.03} & \textbf{78.59} & \textbf{82.78} \\
                                 & Curve     & 35.45 & \textbf{93.85} & 94.43 & \textbf{93.67} \\
            \bottomrule
        \end{tabular}
    \end{threeparttable}
\end{table}

To summarize, the non-repetitive LiDAR demonstrated high Average Precision (AP) performance across the evaluated algorithms, achieving levels comparable to the 128-line repetitive LiDAR. These two systems were generally followed by the 64-line and then the 16-line repetitive LiDARs in terms of AP. However, a key consideration for the non-repetitive LiDAR is its typically smaller HFOV. While it can achieve a very high proportion of quality detections within its operational range (for instance, a 100\% high-AP district proportion in curve scenarios as indicated in Figure \ref{curve_and_crossroad_HQDA}), this limited HFOV means it ultimately covers a smaller absolute number of high-quality detection areas compared to wider HFOV systems like the 128-line and 64-line repetitive LiDARs. PointPillars, PV-RCNN and DSVT all perform similarly with different LiDARs, and are significantly better than PointRCNN.

\section{CONCLUSIONS}

Based on the findings of our experiments, we offer the following recommendations for the deployment of infrastructure-based LiDAR systems:
\begin{itemize}
\item The non-repetitive scanning LiDAR and the 128-line repetitive LiDAR exhibit comparable detection performance across various scenarios, with both consistently outperforming the 64-line system . Furthermore, the non-repetitive LiDAR's detection capability shows a notably slower degradation rate with increasing distance.
\item However, a key trade-off for the non-repetitive LiDAR is its constrained Horizontal Field of View (HFOV), which results in a smaller absolute number of high-quality detection zones compared to wide-angle repetitive systems. Considering its strong performance-to-price ratio, it can be regarded as a highly cost-effective option for specific infrastructure-based deployments.
\item 
Regarding algorithmic performance, pillar-based, voxel-based, and transformer-based 3D object detection methods all yielded commendable AP scores without significant performance disparities among them
\end{itemize}

While this study provides a foundational comparative assessment of repetitive versus non-repetitive LiDAR scanning using CARLA simulations, we acknowledge certain limitations inherent to the simulated environment, such as the incomplete modeling of real-world sensor phenomena like scan distortion, temporal deskewing intricacies, and point cloud intensity data. Our primary intention was to elucidate the fundamental capabilities and trade-offs of these scanning paradigms under standardized conditions. Building on these insights, future investigations will focus on rigorously evaluating the full potential of non-repetitive LiDAR across a broader range of challenging operational scenarios. We anticipate that, with continued advancements in LiDAR manufacturing and signal processing, the promising non-repetitive scanning paradigm will see increasingly widespread adoption in practical applications.

\printbibliography

\begin{IEEEbiography}[
{\includegraphics[width=1in,height=1.25in,clip,keepaspectratio]{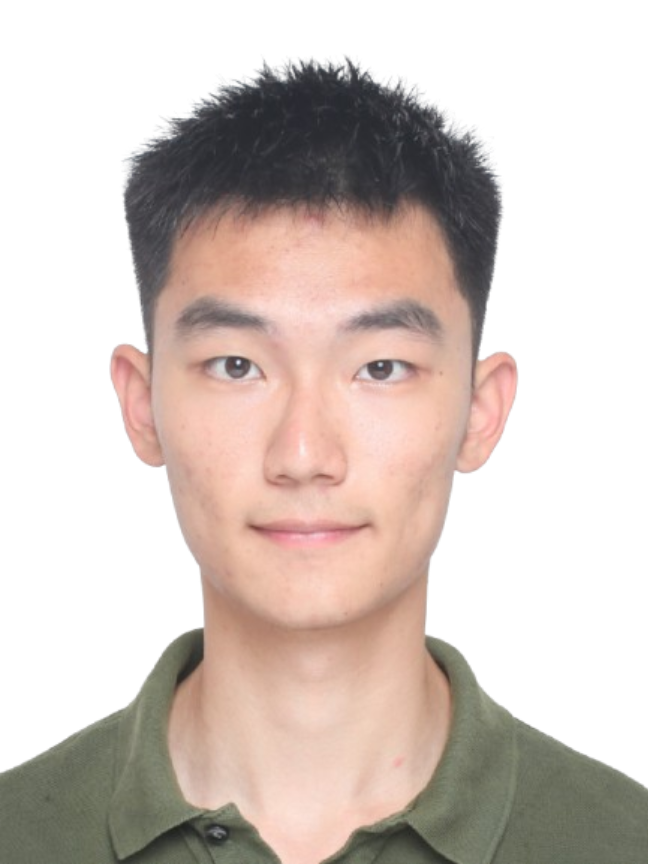}}]
  {Zhiqi Qi}  is pursuing the B.S. degree at Shanghai Jiao Tong University, Shanghai, China. His research interest is in autonomous driving and cooperative driving systems with a focus on sensor-based traffic target detection and LiDAR point cloud analysis. His work aims to contribute to the development of efficient and reliable solutions for enhancing traffic management and safety in urban environments.
\end{IEEEbiography}

\begin{IEEEbiography}[
{\includegraphics[width=1in,height=1.25in,clip,keepaspectratio]{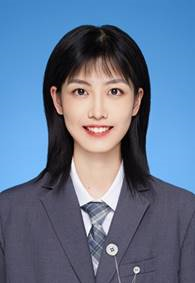}}]
  {RunxinZhao} received the B.Eng. degree from Xi’an Jiao Tong University, Shannxi, China, in 2023. She is currently a Ph.D. student in control science and engineering at the school of automation and intelligent sensing, Shanghai Jiao Tong University, Shanghai, China. Her research interest is in autonomous vehicle localization and point cloud perception.
\end{IEEEbiography}

\begin{IEEEbiography}[
{\includegraphics[width=1in,height=1.25in,clip,keepaspectratio]{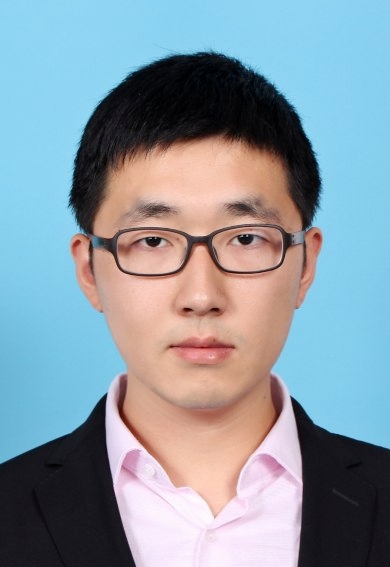}}]
  {Hanyang Zhuang} received the Ph.D. degree from Shanghai Jiao Tong University, Shanghai, China, in 2018. He has worked as a postdoctoral researcher at Shanghai Jiao Tong University from 2020 to 2022. He is currently an assistant research professor at Shanghai Jiao Tong University implementing research works related to intelligent vehicles. His research interest is in autonomous driving and cooperative driving systems.
\end{IEEEbiography}

\begin{IEEEbiography}[
  {\includegraphics[width=1in,height=1.25in,clip,keepaspectratio]{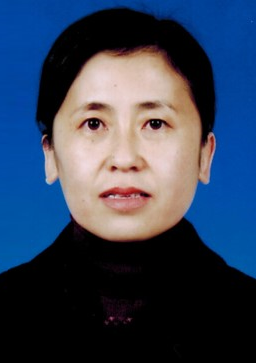}}]
  {Chunxiang Wang} received a Ph.D. degree in mechanical engineering from the Harbin Institute of Technology, China, in 1999. She is currently an Associate Professor at the School of Automation and Intelligent Sensing, Shanghai Jiao Tong University, Shanghai, China. She has been working in the field of intelligent vehicles for more than ten years and participated in several related research projects, such as the European CyberC3 Project and ITER Transfer Cask project. Her research interests include intelligent driving, assistant driving, and mobile robots.
\end{IEEEbiography}

\begin{IEEEbiography}[
  {\includegraphics[width=1in,height=1.25in,clip,keepaspectratio]{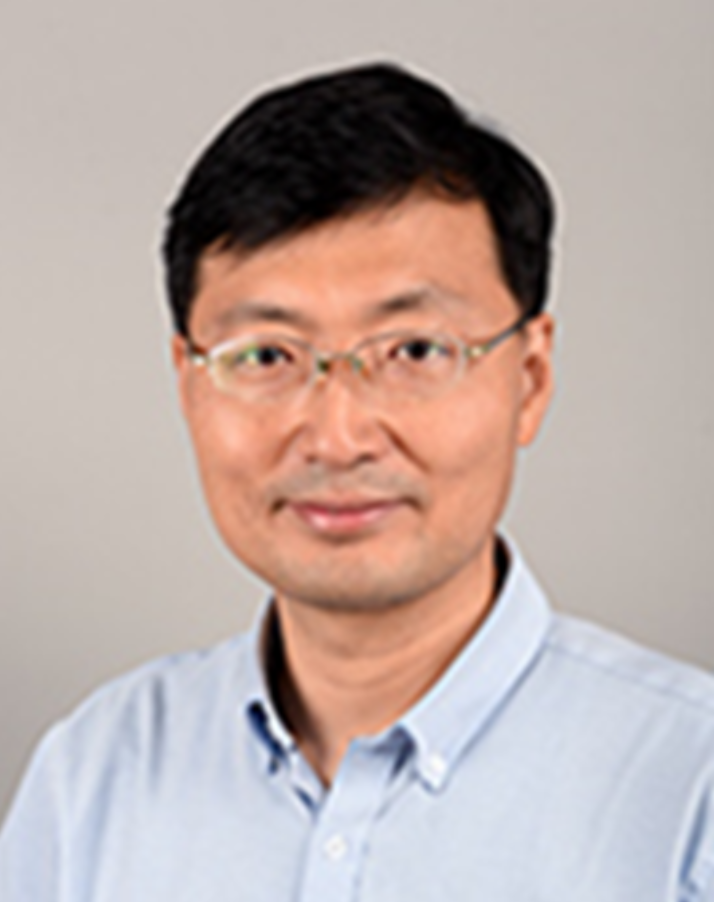}}]
  {Ming Yang} received Master's and Ph.D. degrees from Tsinghua University, Beijing, China, in 1999 and 2003, respectively. He is currently a Full Tenure Professor at Shanghai Jiao Tong University, and the deputy director of the Innovation Center of Intelligent Connected Vehicles. He has been working in the field of intelligent vehicles for more than 20 years. Furthermore, he participated in several related research projects, such as the THMR-V project (first intelligent vehicle in China), European CyberCars and CyberMove projects, CyberC3 project, CyberCars-2 project, ITER transfer cask project, AGV, etc.
\end{IEEEbiography}

\end{CJK}
\end{document}